\documentclass[12pt,a4paper,twoside, openright]{book}
\usepackage{graphicx} 
\usepackage{float}
\usepackage{booktabs} 
\usepackage{subcaption}
\usepackage{url}
\usepackage{cite}
\usepackage{bm}
\usepackage{amsmath}
\usepackage{algorithmic}
\usepackage{amssymb}
\usepackage{textcomp}
\usepackage{xcolor}

\usepackage{setspace}

\usepackage[utf8]{inputenc}
\usepackage[english]{babel}
\usepackage[hidelinks]{hyperref}
\usepackage[capitalize]{cleveref}
\usepackage{mathtools}
\usepackage{fancyhdr}
\usepackage[all]{foreign}
\usepackage{bm}
\usepackage{tikz}
\usepackage{siunitx}
\usepackage{xfrac}

\usepackage[helvratio=0.92]{newtxtext}

\usepackage{glossaries}

\newacronym{ad}{AD}{Annotated Disjunction}
\newacronym{ann}{ANN}{Artificial Neural Network}
\newacronym{ap}{AP}{Access Point}
\newacronym{cfr}{CFR}{Channel Frequency Response}
\newacronym{cmis}{CMISymb}{Conditional Mutual Information on Symbolic data}
\newacronym{cnn}{CNN}{Convolutional Neural Network}
\newacronym{csi}{CSI}{Channel State Information}
\newacronym{cv}{CV}{Computer Vision}
\newacronym{dl}{DL}{Deep Learning}
\newacronym{edl}{EDL}{Evidential Deep Learning}
\newacronym{har}{HAR}{Human Activity Recognition}
\newacronym{kl}{\mbox{KL}}{Kullback–Leibler}
\newacronym{lan}{LAN}{Local-Area Network}
\newacronym{lif}{LIF}{Leaky Integrate-and-Fire}
\newacronym{lpcmci}{LPCMCI}{Latent PCMCI}
\newacronym{lstm}{LSTM}{Long Short-Term Memory}
\newacronym{ltl}{LTL}{Linear Temporal Logic}
\newacronym{mimo}{MIMO}{Multiple-Input Multiple-Output}
\newacronym{mlp}{MLP}{Multi-Layer Perceptron}
\newacronym{mse}{MSE}{Mean Squared Error}
\newacronym{nad}{nAD}{Neural Annotated Disjunction}
\newacronym{nic}{NIC}{Network Interface Card}
\newacronym{nn}{NN}{Neural Network}
\newacronym{ood}{OoD}{Out-of-Distribution}
\newacronym{ofdm}{OFDM}{Orthogonal Frequency-Division Multiplexing}
\newacronym{ofdma}{OFDMA}{Orthogonal Frequency-Division Multiple Access}
\newacronym{pcmci}{PCMCI}{Peter-Clark Momentary Conditional Independence}
\newacronym{phy}{PHY}{Physical Layer}
\newacronym{sdd}{SDD}{Sentential Decision Diagrams}
\newacronym{sdr}{SDR}{Software-Defined Radio}
\newacronym{siso}{SISO}{Single-Input Single-Output}
\newacronym{snn}{SNN}{Spiking Neural Network}
\newacronym{std}{STD}{Standard Deviation}
\newacronym{stdp}{STDP}{Spike-Timing-Dependent Plasticity}
\newacronym{sta}{STA}{station}
\newacronym{vae}{VAE}{Variational Auto-Encoder}
\newacronym{wmc}{WMC}{Weighted Model Counting}
\newacronym{wlan}{WLAN}{wireless Local-Area Network}
\newcommand{\neurospykehar}{\mbox{\textsf{NeuroSpykeHAR}}\xspace}
\newcommand{\templogspyke}{\mbox{\textsf{TempLogSpyke}}\xspace}
\newcommand{\templogcnn}{\mbox{\textsf{TempLogCNN}}\xspace}
\newcommand*\wifi{\mbox{Wi-Fi}\xspace}
\newcommand*\modeli{\mbox{\textbf{Model1}}\xspace}
\newcommand*\modelii{\mbox{\textbf{Model2}}\xspace}
\newcommand*\modeliii{\mbox{\textbf{Model3}}\xspace}
\newcommand{\emdash}{\,---\,}

\begin{document}

\begin{titlepage}
    \begin{figure}
        \centering
        \includegraphics[width=0.5\linewidth]{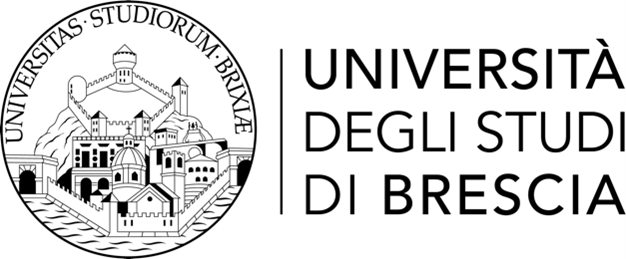}
    \end{figure}
    \begin{center}
        \Large
        DIPARTIMENTO DI INGEGNERIA DELL’INFORMAZIONE\\
        \vspace{0.2cm}
        \Large
        \textit{Corso di Laurea Magistrale in Ingegneria Informatica} \\
        
        \vspace{0.3cm}
        
        \LARGE
        \textbf{Tesi di Laurea}
        
        \vspace{0.2cm}
        
        \LARGE
        \textbf{Approaches to human activity recognition via passive radar}

        \vspace{0.1cm}
        \begin{flushleft}
            \Large
            \textbf{Primo relatore:} \\
            Chiar.mo Prof. Federico Cerutti \\
            
            \textbf{Correlatore:} \\
            Dott. Marco Cominelli

            \hfill \textbf{Laureando:} \\ 
            \hfill Bresciani Christian \\ 
            \hfill Matricola n. 729290
        \end{flushleft}
        
        \vspace{0.5cm}
        
        \small
        Anno Accademico 2023/2024
        
    \end{center}
\end{titlepage}

\selectlanguage{italian}
\glsresetall

\chapter*{Esteso Sommario}
Lo studio analizza metodi innovativi per migliorare l'efficienza e l'adattabilità dei sistemi di riconoscimento delle attività umane (HAR), fondamentali per numerose applicazioni come il monitoraggio sanitario, l'automazione industriale e i sistemi di sicurezza. I sistemi HAR tradizionali, basati su sensori intrusivi come accelerometri o telecamere, sollevano problematiche legate alla privacy. Questa ricerca si concentra su metodi non intrusivi, utilizzando i dati del Channel State Information (CSI) del Wi-Fi, in combinazione con reti neurali avanzate come le Spiking Neural Networks (SNN), che imitano i processi biologici del cervello attraverso eventi discreti detti ``spike.''

La CSI rileva le attività umane analizzando le variazioni del segnale causate dai movimenti nell’ambiente. Questo approccio permette di monitorare le attività senza l’uso di dispositivi indossabili o telecamere, risultando ideale per contesti in cui la privacy è una priorità.
Le CSI misurano le variazioni di intensità del segnale nel tempo e nella frequenza, influenzate da fattori come la presenza e il movimento delle persone. La tesi esplora l’elaborazione dei dati CSI utilizzando reti neurali convoluzionali (CNN) e Spiking Neural Networks (SNN). Le SNN sono particolarmente vantaggiose per HAR grazie alla loro maggiore efficienza, soprattutto se utilizzate con hardware neuromorfico.

Lo studio integra quindi le reti neurali con il ragionamento simbolico, tramite un framework chiamato DeepProbLog. Questo approccio neurosimbolico combina la capacità delle reti neurali di riconoscere schemi con quella dei sistemi simbolici di ragionare logicamente, rendendo il sistema HAR più adattabile. Per allineare i dati CSI ai dati video, da cui viene estratto il ragionamento simbolico, è stata ridotta la frequenza di campionamento dei CSI da 150 a 30 misurazioni al secondo, in linea con la frequenza dei frame video. Tra i diversi metodi di downsampling valutati, la Media è risultata la soluzione più efficace.

L'architettura è stata progettata suddividendo i dati in intervalli temporali, permettendo alla rete di rilevare anche movimenti minimi. Successivamente, i risultati ottenuti dalle reti neurali sono stati integrati in un componente di ragionamento simbolico tramite DeepProbLog, che utilizza regole predefinite basate sui modelli di movimento umano per classificare attività come camminare, correre o applaudire.

Il sistema di ragionamento simbolico classifica le attività utilizzando regole logiche derivate dai dati video. Elementi come l'ampiezza e la velocità dei movimenti degli arti, per parti del corpo come avambracci, braccia e gambe, sono usati per sviluppare regole semplici che distinguono le diverse attività.
Per cogliere la dimensione temporale delle attività umane, la tesi propone, inoltre, l'uso della Logica Temporale, utile per distinguere le attività che comportano una sequenza di movimenti, come camminare o saltare, analizzando come gli elementi identificati variano nel tempo.

I risultati sperimentali confermano l'efficacia dell'approccio proposto. Il modello basato su SNN si è rivelato competitivo rispetto alle tradizionali reti CNN, raggiungendo una precisione del 91,4\% rispetto all'89,1\% della CNN. Un modello neurosimbolico che combina SNN e ragionamento simbolico (NeuroSpykeHAR) ha ottenuto una precisione del 90,45\%, vicino al 94,29\% raggiunto dal modello neurosimbolico più avanzato, DeepProbHAR.

Lo studio dimostra, infine, che è possibile implementare un sistema basato su logiche temporali, proponendo una versione semplificata di tale approccio.

\selectlanguage{english}

\tableofcontents

\listoffigures

\listoftables

\glsresetall
\chapter{Introduction}
\label{chp:Introduction}
\section*{}
This study delves into innovative approaches to enhance the sustainability and adaptability of \gls{har} systems, focusing on non-intrusive sensing methods and advanced neural models. \gls{har} is crucial for a wide range of applications, including healthcare monitoring, smart environments, industrial automation, and security systems.

The aim of this research is to develop a model capable of accurately recognizing human activities using \wifi \gls{csi}, harnessing advanced neural architectures such as \gls{snn} integrated with neurosymbolic reasoning \footnote{Code available at \url{https://github.com/christianbresciani/nscep-spiking-NN}.}. By emphasizing energy efficiency and adaptability, this work addresses several key challenges in current \gls{har} systems.

Traditionally, \gls{har} systems have relied on a range of different sensors, such as accelerometers, cameras, or other physical sensing devices, to capture human activities \cite{chen2012sensor}. While these methods can be effective, they often raise privacy concerns or require individuals to wear intrusive devices \cite{lane2010survey, 9083980}. A more recent and promising alternative is to use \wifi \gls{csi} for \gls{har} \cite{CSI-Sensing, 8735849}. \gls{csi} captures variations in wireless signals caused by human movements, providing an electromagnetic ``signature'' of activity without the need for wearable devices or cameras. This makes it a viable option for monitoring in environments where privacy is paramount, such as healthcare or residential settings.

One of the core components of this thesis is the use of \glspl{snn} \cite{1997SNN} to process \gls{csi} data. \glspl{snn} are neural networks inspired by the way biological neurons operate. Unlike conventional neural networks, which employ continuous activations to represent data, \glspl{snn} rely on discrete events\emdash often referred to as ``spikes''\emdash to transmit information. This makes them a closer approximation of the human brain and provides a more biologically plausible model for neural computation. Importantly, \glspl{snn} offer several advantages that make them particularly suitable for \gls{har}. Firstly, their event-driven nature allows for real-time processing, which is crucial when activities must be detected as they happen \cite{1997SNN}. Real-time capability is an essential requirement for various \gls{har} applications, such as fall detection in elderly care or gesture recognition in human-computer interaction, where quick responses can be life-saving or enable fluid interaction.

Secondly, \glspl{snn} are more energy-efficient compared to traditional deep learning models \cite{roy2019towards}. This energy efficiency can be further enhanced through the use of neuromorphic hardware \cite{davies2018loihi}. Neuromorphic hardware refers to specialised computing platforms, such as Field-Programmable Gate Arrays (FPGAs) and Application-Specific Integrated Circuits (ASICs), that are designed to emulate the functionality of biological neurons and synapses. These platforms support the event-driven computation of SNNs, leading to significant reductions in power consumption and processing time compared to classical von Neumann architectures like CPUs and GPUs. For instance, recent advancements \cite{10.1162/neco_a_01499} have shown that neuromorphic hardware, such as FPGAs, provides the flexibility to implement large-scale \glspl{snn} while maintaining low power usage and high computational speed. This attribute is especially beneficial for deployment on embedded systems and edge devices, where power availability is limited like in a smart home system.

In Chapter \ref{chp:background}, we provide the foundational background for understanding the concepts explored in this thesis, including the fundamentals of \gls{csi}, the application of \glspl{cnn} and \glspl{snn} in \gls{har}, and the principles of neurosymbolic integration. That chapter also establishes the motivation for choosing \wifi-based sensing over other sensor modalities, particularly in terms of privacy and practicality for real-world deployment. Chapter \ref{chp:methodology} outlines the methodology used, detailing the adjustments to the \gls{csi} data handling process, the neural network architectures, and the symbolic components that were implemented. The methodology also describes how temporal logic was incorporated to capture the sequential nature of human activities, ensuring that the recognition process accounts for how events unfold over time, thereby improving model accuracy.

Another significant contribution of this thesis is the introduction of neurosymbolic methods \cite{d2020neurosymbolic} for enhancing the flexibility and interpretability of \gls{har} systems. Neurosymbolic integration combines the feature extraction power of deep learning models with the reasoning capabilities of symbolic systems, which use rules and logical constructs to make decisions. While neural networks excel at recognising patterns from raw data, they often lack transparency, functioning as "black boxes" whose decision-making processes are difficult to interpret. This is a critical limitation in safety-critical applications, such as healthcare monitoring, where understanding how a model reaches a decision is as important as the decision itself. In contrast, symbolic logic offers clear reasoning pathways, making it possible to trace why certain conclusions were drawn.

The neurosymbolic approach employed in this thesis bridges the gap between these two paradigms. By integrating symbolic reasoning with the neural components of our model, we achieve a system that can not only accurately classify activities but also provide traceable decisions. For instance, in a healthcare scenario, the model might indicate that ``the activity is likely a fall because of sudden, high-magnitude fluctuations in movement.'' Embedding declarative knowledge is crucial for user trust and for making the outputs actionable by human supervisors. Moreover, the symbolic layer makes the model flexible to new activities, enabling straightforward updates through the addition of new logical rules without the need for extensive retraining. This adaptability is particularly advantageous in dynamic environments where new activities must be recognised as they emerge, such as in adaptive smart homes or evolving industrial workflows.

We also increased the expressivity of the symbolic component, experimenting with temporal logic as explained in Section \ref{templogic}. Temporal logic \cite{venema2017temporal} allows the model to reason about activities in a way that is consistent with human perception, understanding actions as they occur in a sequence. We integrated it into the neurosymbolic layer, allowing the model to leverage temporal dependencies between different activities and make more informed decisions. For example, activities like walking or jumping can be distinguished based on the timing and sequence of movements, which are represented through logical constructs in temporal logic.

The experimental results presented in Chapter \ref{chp:results} demonstrate the effectiveness of the proposed methods. A significant finding is that the Spiking Neural Network (SNN) performed competitively compared to a traditional Convolutional Neural Network (CNN). Specifically, the \gls{snn} achieved an average accuracy of 91.4\% compared to 89.1\% for the \gls{cnn}, showing its viability as a low-energy, real-time alternative to traditional models. Additionally, this work introduces the first neurosymbolic Spiking Neural Network for \gls{har} (\neurospykehar), which combines the strengths of spiking neurons and symbolic reasoning. The \neurospykehar model achieved an accuracy of 90.45\%, virtually indistinguishable from the existing state-of-the-art neurosymbolic model, DeepProbHAR, which achieved a 94.29\%. These results underscore the effectiveness of combining \glspl{snn} with symbolic reasoning, as the neurosymbolic model provides comparable accuracy while maintaining the benefits of adaptability and lower energy consumption.

\glsresetall
\chapter{Background}
\label{chp:background}
\section*{}
This chapter provides the foundational concepts necessary to understand the research on \gls{har} using \wifi \gls{csi}. It explores how \gls{csi}, a key feature of wireless communication systems, can be leveraged to detect and classify human activities by analyzing signal variations. The chapter delves into different neural network models used for processing \gls{csi} data, including \glspl{cnn} and \glspl{snn}, and introduces the role of neurosymbolic processing in enhancing the interpretability and accuracy of these models. By combining neural and symbolic approaches, this research aims to advance the field of \gls{csi}-based activity recognition.

\glsreset{csi}
\glsreset{cnn}
\glsreset{snn}
\section{Activity Recognition and \wifi}
\label{dataset}
\wifi, and in particular \gls{csi}, can serve as a passive radar to detect human activities.

\gls{csi} is essential in wireless communication, especially in \gls{ofdm} systems such as \wifi, where it estimates the characteristics of the wireless channel. Calculated by the receiver for each incoming \wifi frame, \gls{csi} allows proper equalization of wideband communication channels, mitigating frequency-selective distortions. These distortions, which result from multipath effects in the physical environment, make \gls{csi} analysis crucial for various \wifi sensing applications \cite{OFDM,CSI-Sensing}.

The \gls{csi} effectively captures how wireless signals interact with their surroundings, creating unique interference patterns influenced by factors such as room layout, furniture arrangement, and human presence and movement. 
Essentially, the \gls{csi} acts as an electromagnetic signature of the environment, with \wifi receivers functioning similarly to passive radars under suitable conditions. 
This capability makes \gls{csi}-based \gls{har} possible, as the variations in signal characteristics\emdash due to human movement\emdash can be correlated with specific activities \cite{CSI-Sensing}.

\begin{figure}[ht] \centering \includegraphics[width=0.6\textwidth]{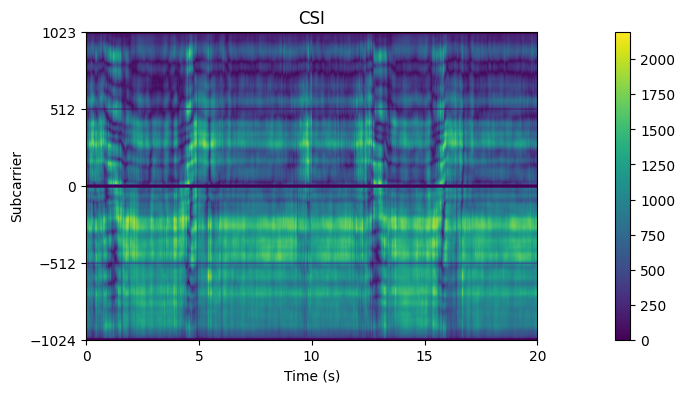} \caption{Illustrative figure showing a snippet of \gls{csi} data captured by one antenna while a person walks.} \label{csi} \end{figure}

Figure \ref{csi} illustrates a segment of data captured by a single antenna, depicting the \gls{csi} magnitude while a person walks. As the individual moves around the room, the signal's characteristics fluctuate due to varying scattering effects on the human body. This variation is visualised in a spectrogram, illustrating changes in signal intensity across time and frequency. The fundamental premise of \gls{csi}-based \gls{har} is the ability to correlate these variations with specific human activities.

Our research utilises a publicly available \gls{csi} dataset.\footnote{\url{https://github.com/ansresearch/exposing-the-csi} on 4 September 2024} The experimental setup for this dataset includes two Asus RT-AX86U devices positioned on opposite sides of a room measuring approximately 46 square metres. One device continuously generates dummy IEEE 802.11ax (\wifi 6) traffic at a rate of 150 frames per second using a frame injection feature \cite{CSI-Extraction}. The other device, acting as a monitor, receives these \wifi frames and records the corresponding \gls{csi} for each of its four antennas. During this process, a participant performs various activities in the centre of the room.

\begin{figure}
    \centering
    \includegraphics[width=0.95\columnwidth]{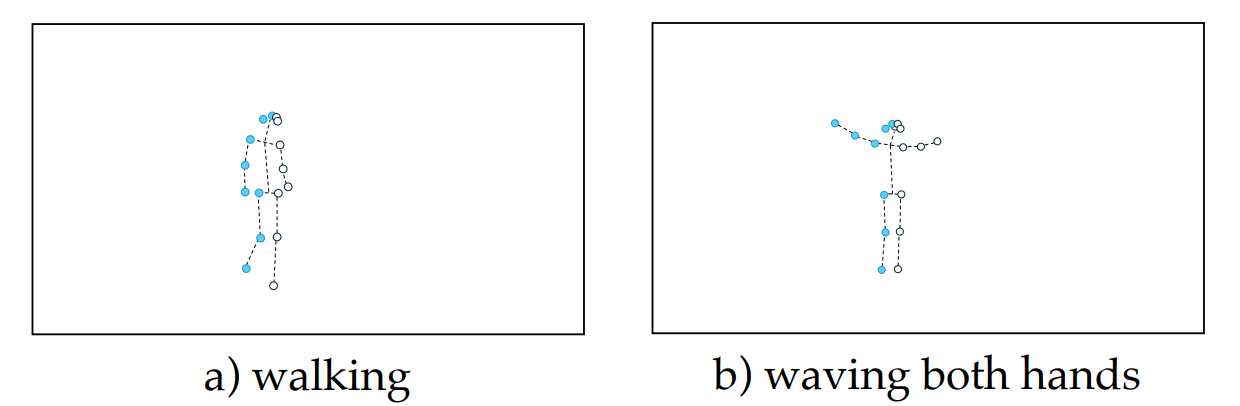}
    \caption{Sample of the video dataset for two different activities: a) {\it walking} and b) {\it waving both hands}. The key points in every video frame help to discern the right side (highlighted with coloured dots) from the left side of the candidate. Image sourced from \cite{cominelliFusion2024} with the author’s permission.}
    \label{fig:videopose3d}
\end{figure}

The \gls{csi} data is roughly synchronised with a video recording of the activities, captured using a fixed smartphone camera. To anonymise the participants, 
 VideoPose3D~\cite{VideoPose3D} is used to extract a model of the individual's movements, identifying 17 key points that track the motion of major joints Figure \ref{fig:videopose3d}. These key points are saved as (x, y) coordinates for each video frame. Although the dataset includes \gls{csi} data for twelve different activities, only seven activities have both \gls{csi} and video data: walking, running, jumping, squatting, waving both hands, clapping, and wiping. Each activity provides 80 seconds of \gls{csi} data sampled at 150 \gls{csi} samples per second and corresponding video data sampled at 30 frames per second.

There are two primary approaches to activity recognition: declarative and data-driven. Declarative approaches provide classification rules to define the activity, such as describing running as the rapid, alternating action of pushing off and landing on the ground with one's feet \cite{storf2009rule,theekakul2011rule,atzmueller2018explicative}. However, these approaches are limited in their ability to handle unstructured data such as \wifi \gls{csi}, and we are not aware of any declarative methods applied to \gls{csi}-based \gls{har}.

In contrast, data-driven approaches are designed to handle complex data types for which direct rule definition is challenging \cite{meneghello2022,bahadori2022rewis,liu2020,fusion2023}. These methods, though more opaque and less flexible than declarative approaches, are well-suited to our needs. Several \gls{har} systems work by deriving a physically-related quantity from sensors like \gls{csi} and then using this data to train a deep learning classification system. 

In previous works a principled approach to \gls{har} using a \gls{vae} generative model to compress the sensors' information and then derive from the compressed data the activity with a data-driven only approach \cite{fusion2023} and with a data-driven plus declarative approach \cite{cominelliFusion2024}.

\section{Neural CSI Processing}
\label{networks}
In this section, we delve into the two primary types of neural networks relevant to process \gls{csi} data: \glspl{cnn} and \glspl{snn}. While \glspl{cnn} are widely recognised as a standard benchmark for tasks involving spatial data such as images, in this work, we focus on \glspl{snn} as our primary network model, using \glspl{cnn} as a comparative baseline. The following sections provide a detailed overview of each network type, highlighting their structures, functionalities, and relevance to our study.

\subsection{Convolutional Neural Networks}
\glspl{cnn} are a powerful class of deep learning models that have become integral to the field of machine learning, particularly for tasks involving data with a grid-like structure, such as images, audio spectrograms, and even \gls{csi} in wireless communications \cite{NIPS2012_c399862d, 7952132, 9322323}.

\glspl{cnn} are fundamentally different from traditional feedforward neural networks in that they are specifically designed to process and learn from spatial data. Traditional neural networks treat input data as a one-dimensional vector, disregarding the spatial relationships between elements. In contrast, \glspl{cnn} preserve and exploit the spatial structure of the data by using convolutional operations that allow them to recognise patterns such as edges, textures, and shapes in images, or temporal and frequency patterns in \gls{csi} data.

The convolutional layer \cite{726791} is the primary building block of a CNN, and it is this layer that gives the network its name. In a convolutional layer, a set of learnable filters, also known as kernels, is applied to the input data. Each filter slides across the input, performing a convolution operation, which involves computing the dot product between the filter and a small region of the input data. This operation is repeated across the entire input, generating a feature map that highlights specific features of the data, such as edges in an image or temporal changes in a \gls{csi} signal.

One of the key advantages of using convolutional layers is the concept of parameter sharing. Unlike in fully connected layers, where each neuron has its own set of weights, in convolutional layers, the same set of weights (i.e., the filter) is used across different regions of the input. This significantly reduces the number of parameters that need to be learned, making the network more efficient and less prone to overfitting. Additionally, the use of convolutional layers allows \glspl{cnn} to be invariant to small translations of the input, meaning that they can recognise patterns regardless of where they appear in the data.

After the convolution operation, the feature maps are passed through an activation function to introduce non-linearity into the network. Non-linearity is crucial because it enables the network to learn complex, non-linear relationships in the data, which are essential for capturing the intricacies of real-world patterns. The most commonly used activation function in \glspl{cnn} is the Rectified Linear Unit (ReLU) \cite{nair2010rectified}, which applies the following simple operation: any negative value in the feature map is replaced with zero, while positive values remain unchanged.

The ReLU activation function is particularly popular due to its simplicity and effectiveness. It helps prevent the vanishing gradient problem \cite{pmlr-v15-glorot11a}, which can occur with other activation functions like the sigmoid or hyperbolic tangent, where gradients become very small and slow down the learning process. By maintaining positive linearity, ReLU ensures that the network can continue to learn efficiently as the depth of the network increases, which is often necessary for capturing more abstract features in the data.

Pooling layers \cite{boureau2010theoretical} are another crucial component of \glspl{cnn}, serving the dual purpose of reducing the spatial dimensions of the feature maps and controlling overfitting by summarising the presence of features in local regions. The most common type of pooling operation is max pooling \cite{1467551}, which extracts the maximum value from a small patch of each feature map. This operation not only reduces the dimensionality of the data, making subsequent layers more computationally efficient, but it also helps the network become invariant to small distortions and translations in the input data.

For example, in image data, max pooling can summarise the presence of edges or textures in a local region, regardless of their exact position within that region. In the context of \gls{csi} data, pooling layers can summarise the temporal and frequency variations in the signal, capturing the most significant changes that correspond to different human activities. This process of dimensionality reduction is crucial in deep networks, where the number of parameters can quickly become overwhelming if not carefully managed.

After several convolutional and pooling layers, the output feature maps are typically flattened into a one-dimensional vector, which is then fed into one or more fully connected layers. These layers are similar to those found in traditional neural networks, where each neuron is connected to every neuron in the previous layer. Fully connected layers are responsible for high-level reasoning in the network, as they combine the features extracted by the convolutional layers to make a final prediction.

The fully connected layers are where the network performs classification. In the case of image recognition, for example, the output of the final fully connected layer might be a vector representing the probabilities of different classes (e.g., cat, dog, car). In our work, the fully connected layers take the high-level features extracted from the \gls{csi} data and predict the most likely human activity being performed, such as walking, running, or waving hands.

While the convolutional and pooling layers are responsible for feature extraction, the fully connected layers integrate these features to produce the final output. The weights in the all both the convolutional layers and fully connected layers are learned through backpropagation, where the network adjusts its parameters based on the difference between the predicted output and the true labels during training.

\subsection{Spiking Neural Networks}
\glspl{snn} are a class of artificial neural networks designed to more closely replicate the behavior and processing mechanisms of biological neural systems \cite{1997SNN}. Unlike conventional \glspl{ann}, which typically rely on continuous activation functions and use gradient-based learning methods like backpropagation, \glspl{snn} process information using discrete events known as spikes. These spikes are akin to the action potentials or pulses seen in biological neurons, enabling \glspl{snn} to encode information in both the timing and frequency of these events. This approach offers several advantages, including the ability to capture temporal dynamics, perform sparse and energy-efficient processing, and integrate into neuromorphic hardware systems \cite{brainsci12070863}.

The basic architecture of an \gls{snn} consists of several layers, each comprising spiking neurons connected through synapses. The structure is analogous to traditional \glspl{ann}, but the way data is represented and propagated through the network is fundamentally different \cite{lv2024efficienteffectivetimeseriesforecasting}.

The input layer is responsible for converting external stimuli, such as sensory data or encoded signals, into spike trains. These spike trains are sequences of discrete spikes that represent the input signal's intensity and temporal characteristics. Different encoding methods can be employed in this layer. One common method is rate coding, where the intensity of the input signal is encoded by the frequency of spikes\emdash higher input values result in a higher spike frequency. Another method is temporal coding, where the timing of each spike carries critical information about the input signal, allowing the network to encode time-dependent patterns more effectively.

The hidden layers in an \gls{snn} are where most of the computation occurs. Each layer consists of spiking neurons that process the spike trains received from the previous layer. A widely used model for spiking neurons is the \gls{lif} model. In the \gls{lif}  model, each neuron integrates incoming spikes over time, resulting in a change in its membrane potential, 
\( U[t] \). This potential evolves according to the following equation:

\[
U[t+1] = \beta U[t] + I_{\rm in}[t+1] - R
\].

Here, 
\( \beta \) is a decay factor representing the leakage of the membrane potential over time, 
\( I_{\rm in}[t] \) is the input current at time 
\( t \), and \( R \) is the reset value applied after the neuron fires a spike. When the membrane potential exceeds 
a certain threshold \( U_{\rm thr} \), the neuron
generates an output spike, 
\( S[t+1] \), as given by:

\[
U[t] > U_{\rm thr} \Rightarrow S[t+1] = 1
\].

After firing a spike, the membrane potential is reset according to the selected reset mechanism, either by subtracting the threshold (Subtract Reset) or resetting the potential to a lower value (Zero Reset). The specific behavior of these neurons allows \glspl{snn} to capture and process temporal information across multiple layers, making them particularly suited for dynamic and time-dependent tasks.

The output layer in an \gls{snn} translates the spike trains from the hidden layers into a final decision or output. This can be achieved through various decoding strategies. One approach is to count the number of spikes generated by each neuron in the output layer over a specific time window, with the neuron having the highest count determining the output. Another method involves analyzing the precise timing of spikes to extract more nuanced information. This output decoding process is crucial for interpreting the network's response in a meaningful way, especially in applications such as classification, pattern recognition, and decision-making.

The dynamics of spiking neurons in \glspl{snn} are governed by models like the \gls{lif}, which simulates the electrical properties of biological neurons. The \gls{lif}  model, while relatively simple, captures essential aspects of neuronal behavior, such as membrane potential integration, leakage, and the generation of spikes. The membrane potential 
\( U[t] \) is a key variable, representing the neuron's internal state at time \( t \).
It is influenced by incoming spikes (input current) and the inherent properties of the neuron, such as its decay rate (\( \beta \)) and threshold potential (\( U_{\rm thr} \)).

Learning in \glspl{snn} is achieved through backpropagation using surrugate functions so to have a gradient that can be used in the backward pass. 
The package SNNTorch\footnote{\url{https://github.com/jeshraghian/snntorch} on 10 September 2024} \cite{eshraghian2021training} we use substitutes 
\[\begin{split} S=\begin{cases} 1 & \text{if U} > \text{U$_{\rm thr}$} \\
0 & \text{if U} < \text{U$_{\rm thr}$} \end{cases}\end{split}\]
with 
\[\begin{split}S&\thickapprox\frac{1}{\pi}\text{arctan}(\pi U )\end{split}\].

\glspl{snn} offer several advantages over traditional \glspl{ann}, particularly in terms of energy efficiency and temporal processing capabilities \cite{9927729}. The event-driven nature of \glspl{snn} means that neurons only consume energy when they spike, which can lead to significant energy savings, especially in large-scale networks. This property is particularly valuable in neuromorphic hardware implementations, where power consumption is a critical concern.

Moreover, the ability of \glspl{snn} to naturally encode and process temporal information makes them ideal for applications involving time-series data or tasks that require real-time processing. Examples include speech and audio recognition, video analysis, robotic control, and even brain-machine interfaces. The temporal coding and learning mechanisms allow \glspl{snn} to excel in scenarios where traditional \glspl{ann} might struggle, particularly when dealing with continuous streams of dynamic, time-varying data.

\subsection*{}
In summary, while \glspl{cnn} have established themselves as a powerful tool for processing spatial data and have been used effectively in activity recognition tasks, \glspl{snn} offer a more biologically plausible and energy-efficient alternative, particularly when dealing with temporal information. In this work, we utilise \glspl{snn} as the primary network model for \gls{csi}-based activity recognition, with \glspl{cnn} serving as a standard for comparison. The exploration of \glspl{snn} for this specific task is the first step of this thesis work, as we seek to demonstrate their potential advantages in accurately and efficiently recognising human activities through wireless signals.

\section{Neurosymbolic CSI Processing}
\label{logic}
Neurosymbolic learning and reasoning is an emerging paradigm that combines the strengths of symbolic logic-based reasoning with the perceptual capabilities of neural networks. This integration is essential for addressing complex problems where both structured reasoning and pattern recognition are required. In the context of activity recognition using \gls{csi} data, neurosymbolic approaches allow us to extract relevant features from noisy, high-dimensional data through neural networks, while simultaneously using logic to reason about these features in a transparent and interpretable manner. This is particularly important for tasks where explainability, flexibility, and handling uncertainty are crucial, as it enables a deeper understanding of the underlying processes.

In this section, we first explore DeepProbLog, a key tool in our neurosymbolic approach. DeepProbLog builds upon ProbLog, a foundational framework for probabilistic logic programming. Since ProbLog is the core of DeepProbLog, it is essential to first introduce ProbLog before delving into the specifics of DeepProbLog.

\subsection{Problog}
ProbLog \cite{FIERENS_VAN, de2007problog} is a probabilistic logic programming language that merges the declarative power of logic programming with the ability to model uncertainty through probability theory. At its core, ProbLog allows users to define probabilistic facts—statements that have associated probabilities, reflecting the inherent uncertainty of real-world scenarios. 

For instance, a probabilistic fact in a ProbLog program might state
\begin{verbatim}
    0.2::earthquake.
\end{verbatim}
indicating a 20\% probability of an earthquake occurring.

These probabilistic facts are combined with logical rules, which define relationships between different facts. A typical rule might look like
\begin{verbatim}
    alarm :- earthquake.
\end{verbatim}
signifying that the alarm will sound if an earthquake occurs. The true power of ProbLog lies in its ability to handle queries about the likelihood of certain events or states, given the defined probabilistic facts and logical rules. For example, one could query the probability that an alarm will sound, considering all possible causes, such as earthquakes and burglaries.

Technically, ProbLog operates by first grounding the logical program transforming it into a fully instantiated form where all variables are replaced by constants. This process results in a propositional representation of the program. ProbLog then uses \gls{wmc} \cite{renkens2014explanation, sang2005performing}, a technique that counts the number of models (possible worlds) that satisfy the query, weighted by the probabilities of the contributing facts. To manage the complexity of \gls{wmc}, ProbLog employs \glspl{sdd} \cite{vlasselaer2014compiling}, which are compact representations of the propositional formulae. \glspl{sdd} enable efficient computation of probabilities by reducing the logical formula to a structure where \gls{wmc} can be performed efficiently.

ProbLog's ability to handle a wide range of probabilistic scenarios, from simple binary events to complex multi-valued logic, makes it an invaluable tool for applications in areas such as risk assessment, decision support systems, and automated reasoning under uncertainty. The declarative nature of logic programming in ProbLog also means that complex probabilistic models can be expressed concisely and interpreted easily, which is crucial for applications that require transparency and explainability.

\subsection{DeepProbLog}
DeepProbLog \cite{NEURIPS2018_dc5d637e} is an extension of ProbLog, designed to integrate the pattern recognition capabilities of neural networks with the logical reasoning framework of probabilistic logic programming. This integration is achieved through the introduction of \textit{neural predicates} and \textit{\glspl{nad}}, which allow neural networks to directly influence the probabilities of certain outcomes in a logic program.

Neural predicates in DeepProbLog are a powerful concept where a neural network determines the probability of a logical predicate being true. For example, if we have a neural predicate \texttt{digit(X, D)}, the network processes an image \texttt{X} and outputs a probability distribution over possible digits \texttt{D}. This distribution reflects the network's confidence in each digit being the correct label for the image. This setup allows DeepProbLog to seamlessly combine the data-driven, subsymbolic learning of neural networks with the rule-based, symbolic reasoning of logic programming.

The concept of \glspl{nad} takes this integration further by allowing the output of a neural network to define the probability distribution over multiple possible outcomes. 
An nAD might look like 
\begin{verbatim}
    nn(m\_digit, [X], Y, [0, ..., 9]) :: digit(X,Y).
\end{verbatim}
where \texttt{m\_digit} is a neural network model that takes an input image \texttt{X} and outputs probabilities for each digit from 0 to 9. The probabilities assigned to each possible digit by the neural network are then used in the logic program to infer higher-level outcomes.

DeepProbLog retains the essential features of ProbLog, such as its inference mechanism and the ability to handle probabilistic logic programs, while adding the capability to integrate neural networks as probabilistic components. This integration allows DeepProbLog to tackle tasks that require both perception (handled by neural networks) and reasoning (handled by the logic program). For example, in the task of handwritten digit recognition and arithmetic, DeepProbLog can use a neural network to recognize individual digits and then use logical rules to perform arithmetic operations on those digits. The result is a probabilistic logic program that can reason about complex scenarios involving both symbolic and subsymbolic data.

The training process in DeepProbLog is another area where it diverges from traditional ProbLog. Training in DeepProbLog involves optimizing both the parameters of the neural networks and the probabilistic parameters of the logic program. This is done using gradient-based learning, where the gradients are computed for both the probabilistic logic part and the neural network part of the model. The backpropagation algorithm, traditionally used in neural network training, is extended to handle the probabilistic logic components, allowing for end-to-end learning across both subsymbolic and symbolic layers of the model.

One of the key advantages of DeepProbLog is its ability to perform \textit{end-to-end learning} where the entire model, including both the neural networks and the logic program, is trained simultaneously based on the final output of the model. This is particularly useful in scenarios where the logic program provides a high-level structure or constraints that guide the learning process of the neural networks. For instance, in a task that involves recognizing objects in images and then reasoning about their relationships, DeepProbLog can ensure that the neural network's output is consistent with the logical constraints, leading to more accurate and interpretable models.

Furthermore, DeepProbLog’s ability to incorporate neural networks opens up new possibilities for combining deep learning with structured, interpretable reasoning. By integrating probabilistic logic with neural networks, DeepProbLog can model a wide range of complex problems that require both detailed data analysis and high-level reasoning, such as visual question answering, where a model needs to understand and reason about visual data, or activity recognition from sensor data, where temporal and spatial patterns must be recognized and interpreted logically.

\subsection*{}
In this thesis, DeepProbLog plays a central role in bridging the gap between deep learning and logical reasoning, offering a powerful framework for tackling complex tasks that require both capabilities. The integration of symbolic logic with neural networks in DeepProbLog allows us to model and solve problems that are beyond the reach of traditional deep learning models or pure logic-based systems.

The application of DeepProbLog in our research is particularly focused on activity recognition using \gls{csi} data. Here, the neural networks within DeepProbLog learn to extract relevant features from raw \gls{csi} data, while the probabilistic logic component reasons about these features to infer the most likely human activities. This approach leverages the strengths of both deep learning and logic programming, providing a robust and interpretable model for complex real-world tasks.

The training process in DeepProbLog represents the second and third steps of this thesis work. As second step, similarly to \cite{cominelliFusion2024}, we employ DeepProbLog with a simple logic to demonstrate the effectiveness of neuro-symbolic integration with \glspl{snn} for \gls{csi}-based activity recognition. The third step involves an advanced application of DeepProbLog using temporal logic, as detailed in Section \ref{templogic}.

\glsresetall
\section*{Summary}
Chapter \ref{chp:background} provides essential insights into \gls{har} using Wi-Fi \gls{csi}, highlighting \gls{csi}'s role in detecting human activities by analyzing signal changes caused by movement in the environment. It then delves into neural processing techniques, focusing on \glspl{cnn} for spatial pattern recognition and \glspl{snn} for energy-efficient temporal processing, comparing the strengths of both. Finally, the chapter introduces neurosymbolic \gls{csi} processing, which merges neural networks with symbolic logic, offering a powerful, interpretable approach to extracting and reasoning about complex data, using tools like ProbLog and DeepProbLog for enhanced decision-making in \gls{har}.

\glsresetall
\chapter{Methodology}
\label{chp:methodology}
\section*{}
In this chapter, the methodology for integrating \wifi \gls{csi} with \glspl{snn} and symbolic reasoning is outlined. Building upon existing research, this section describes how \gls{csi} data is processed to enhance \gls{har}. Key adjustments are made to handle \gls{csi} data effectively, modify the neural network architecture, and integrate symbolic reasoning with temporal logic. These modifications aim to improve the model's accuracy and transparency, providing a comprehensive framework for fusing neural and symbolic techniques to classify human activities.

\glsresetall
\section{Existing Methodology}
\label{fusion2024}
This research builds directly upon the methodological innovations presented in the DeepProbHAR \cite{cominelliFusion2024}. The study developed a neuro-symbolic architecture aimed at improving the transparency and accuracy of \gls{har} using \wifi \gls{csi} data. Traditional data-driven approaches to \gls{har}, while effective, often suffer from opacity and require substantial amounts of labelled data. DeepProbHAR addresses these limitations by integrating symbolic reasoning with neural network-based processing, allowing for more interpretable and flexible models.

The technical foundation of DeepProbHAR is the combination of symbolic AI, which provides explicit, rule-based reasoning, with the pattern recognition capabilities of neural networks. The symbolic component is derived from the anonymised video recording of the activities, where human activities are described through a set of rules based on the movement of specific body parts. These rules are then applied to classify activities detected through the \wifi \gls{csi} data. The \gls{csi} data captures how electromagnetic waves interact with the human body as it moves, providing detailed but complex signals that can be correlated with different human activities.

To manage the high dimensionality and complexity of \gls{csi} data, the study employed \glspl{vae} \cite{kingma2013auto}. \glspl{vae} are a class of generative models that encode input data into a compressed latent space while preserving the essential features needed for classification tasks. Specifically, the \gls{vae} used in DeepProbHAR maps short sequences of \gls{csi} data, which are sampled using a sliding window technique, onto a latent space characterized by a bivariate Gaussian distribution. This latent space is defined by four parameters: two means and two variances along orthogonal axes, which effectively summarize the key characteristics of the \gls{csi} data.

The input to the \gls{vae} is structured as a tensor of size (W × S × A), where:
\begin{itemize}
    \item \textbf{W} is the number of \gls{csi} samples within a defined time window (fixed at 3 seconds, corresponding to 450 samples),

    \item \textbf{S} is the number of subcarriers in a \wifi frame (2048 subcarriers for 160-MHz 802.11ax \wifi frames),

    \item \textbf{A} is the number of antennas (ranging from 1 to 4 depending on the architecture).
\end{itemize}
This encoding process significantly reduces the dimensionality of the \gls{csi} data while retaining the information necessary for accurate activity classification.

Once the \gls{csi} data is encoded into the latent space by the \gls{vae}, it is fed into a series of \glspl{mlp} \cite{rosenblatt1958perceptron}. These \glspl{mlp} are trained to classify the human activities by recognizing patterns in the compressed \gls{csi} data. Each \gls{mlp} is designed to correspond to specific decision nodes derived from the symbolic rules encoded from the video data. In DeepProbHAR, six \glspl{mlp} are used, each responsible for detecting different features related to limb movements, such as whether the right lower leg moves or whether the left arm is active.

Each \gls{mlp} in the system has two hidden layers, with 8 neurons per layer, activated by a ReLU function, and a binary output layer with a SoftMax activation function \cite{bridle1990probabilistic}. The binary outputs of these \glspl{mlp} correspond to simple movement features, which are then combined using logic rules to determine the overall activity. The architecture of DeepProbHAR ensures that these neural classifiers are tightly integrated with the symbolic reasoning process, making the system’s decisions more interpretable.

The symbolic component of DeepProbHAR is based on declarative knowledge extracted from a video dataset. This dataset was used to track 17 key points on the human body, corresponding to major joints like shoulders, elbows, knees, and ankles. From these points, the angles and movements of various limbs were calculated using basic trigonometric functions. For instance, the angle of the right upper arm was determined by calculating the \(\arctan\) of the difference in y-coordinates over the difference in x-coordinates between the shoulder and elbow.

These limb movements were then quantified into features representing the dynamic range of motion within each time window. A tree was constructed from these features, where each node in the tree represented a binary decision based on whether a specific limb movement exceeded a certain threshold. This tree served as the foundation for the symbolic reasoning in DeepProbHAR, guiding the classification process by linking specific combinations of limb movements to particular activities.

The study also investigated various strategies for fusing data from multiple antennas, given that the devices employed can collect \gls{csi} data from four different antennas. Three fusion strategies were explored:
\begin{enumerate}
    \item \textbf{No-Fusion}: \gls{csi} data from individual antennas were processed independently using separate \glspl{vae} and \glspl{mlp} (VAE input with A=1). 
    
    \item \textbf{Early Fusion}: \gls{csi} data from all antennas were stacked together into a single tensor, which was then processed collectively by the \gls{vae} (VAE input with A=4).

    \item \textbf{Delayed Fusion}: The \gls{csi} data from each antenna were processed separately by individual \glspl{vae}, and the resulting latent space parameters were concatenated before being fed into the \glspl{mlp} for classification (VAE input with A=1).
\end{enumerate}

The results indicated that delayed fusion, which maintains the separation of data until the final classification stage, provided the best accuracy. This approach benefits from the independent processing of each antenna's data, capturing more detailed and distinct features from different perspectives.

Experimental validation was conducted using a publicly available dataset comprising both \gls{csi} data and anonymized video recordings. The system was trained on 80\% of the dataset and tested on the remaining 20\%. The results demonstrated that DeepProbHAR could achieve accuracy levels comparable to state-of-the-art \gls{har} models while maintaining greater interpretability due to its neuro-symbolic architecture. The confusion matrices generated during testing highlighted the effectiveness of the delayed fusion strategy, particularly in distinguishing between complex activities such as running and walking.

\section{Methodology Adjustments}
\label{methodologyAdj}
\subsection{Changes to CSI Data Handling and Downsampling}
One of the key modifications introduced in this work, compared to the original DeepProbHAR architecture, involves the preprocessing of \gls{csi} data. In the original study, \gls{csi} data were sampled at a rate of 150 measurements per second, which was significantly higher than the 30 frames per second (fps) used in the corresponding video recordings. This difference in temporal resolution created challenges, particularly as human-like abilities needed to be considered to align the data with how humans perceive and process information. To address this, we opted to downsample the \gls{csi} dataset by a factor of five, reducing the sampling rate from 150 \gls{csi} measurements per second to 30 \gls{csi} measurements per second, effectively matching the video frame rate.

We explored four different downsampling strategies to determine the most effective method for reducing the data while preserving its integrity. The techniques were as follows:

\begin{itemize}
    \item Mean Downsampling: This approach calculates the mean value of every 5 consecutive \gls{csi} measurements, providing a smooth average that represents the overall trend within each group.

    \item Median Downsampling: Here, the median value of each group of 5 consecutive \gls{csi} measurements is selected. The median is less sensitive to outliers than the mean, making it more robust in cases where occasional large fluctuations occur in the data.

    \item Minimum Downsampling: This method takes the minimum value from each group of 5 \gls{csi} measurements. It is useful for capturing the lower bound of the signal fluctuations, but can result in data loss if the minimum value is not representative of the overall signal trend.

    \item Maximum Downsampling: This technique selects the maximum value from each group of 5 \gls{csi} measurements, capturing the upper bound of the signal within that interval. Like minimum downsampling, this method can discard information about smaller fluctuations within the group.
\end{itemize}

We considered only the magnitude of the \gls{csi} ignoring the phase information as it is very noisy also \cite{cominelliFusion2024} used only the magnitude.

To evaluate the effectiveness of each downsampling method and ensure that the reduced dataset still captured the key features of the original \gls{csi} data, a series of statistical and visual evaluations were conducted. These evaluations were designed to assess the similarity between the downsampled data and the original data across different summarisation metrics.

The primary metrics used for this evaluation were:
\begin{enumerate}
    \item \gls{mse} of Mean: This metric calculates the \gls{mse} between the mean values of the original \gls{csi} dataset and the downsampled dataset. The aim was to assess how well the downsampled data preserved the overall trend and central tendency of the original signal.

    \item \gls{mse} of \gls{std}: This evaluates the \gls{mse} between the standard deviation of the original and downsampled datasets, providing insight into how well the downsampled data captured the variability in the signal over time.

    \item Norm1 of Differences: This metric sums the absolute differences between consecutive \gls{csi} measurements, both in the original and downsampled datasets. It measures how accurately the downsampled data keeps the temporal dynamics of the original signal, particularly the degree of fluctuation between adjacent measurements.    
\end{enumerate}

These evaluations were performed in two configurations:

\begin{itemize}
    \item Full Time Frame Evaluation: The metrics were calculated over the entire dataset for each activity. This provided a global view of how well the downsampled data aligned with the original data across long time periods.

    \item Time-Windowed Evaluation: To capture short-term dynamics, the same metrics were computed within 3-second time windows (450 \gls{csi} measurements in the original dataset, 90 \gls{csi} measurements after downsampling). 
\end{itemize}

This time-windowed analysis allowed us to assess how well the downsampling preserved the finer details of the signal, which are critical for recognising activities that involve rapid or subtle movements.

In addition to these statistical evaluations, we performed a visual inspection of the data in the frequency domain by computing the Fourier transform of the original and downsampled \gls{csi} signals. This step was critical in determining whether the downsampling process affected the spectral characteristics of the signal, which are important for identifying periodic patterns associated with different activities (e.g., walking, running).

\begin{table}[htbp]
    \centering
    \begin{tabular}{lrrr}
        \toprule
        \textbf{Mode} & \textbf{MSE(E[x])} & \textbf{MSE(\(\sum\| \mathbf{x}_i - \mathbf{x}_{i+1} \|_1\))} \\
        \midrule
        Max & 9.99e+03 & 372.20 \\
        Mean & 1.11e-26 & 160.56 \\
        Median & 2.22e+01 & 406.67 \\
        Min & 1.14e+04 & 822.91 \\
        \bottomrule
    \end{tabular}
    \caption{Table showing the full time evaluation of MSE\_Mean and MSE\_Diff\_Norm for different downsampling modes for the \textit{Walk} action.}
    \label{tab:mse}
\end{table}

After evaluating the performance of the different downsampling methods, it was determined that Mean Downsampling provided the best balance between preserving the overall characteristics of the original signal and reducing data volume. As shown in Table \ref{tab:mse}, the mean undersampling method maintained the closest match in terms of \gls{mse} for the mean values and the Norm1 of differences. Its overall performance across different metrics and activities made it the most suitable choice for further experiments.

\begin{figure}[htbp]
    \centering
    \includegraphics[width=1\linewidth]{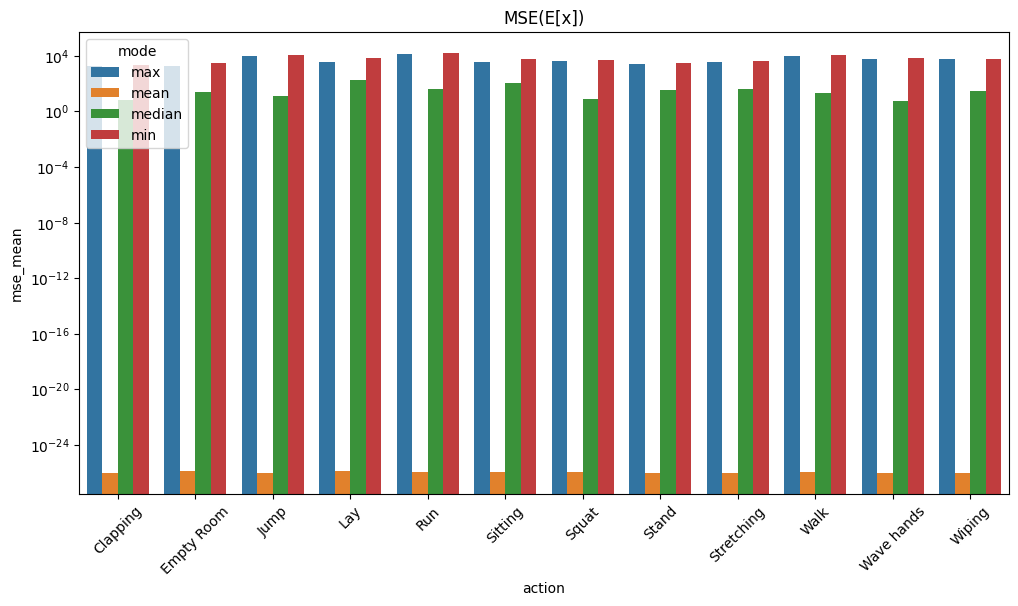}
    \caption{Figure showing the MSE over the frequencies between the mean of the original data and the downsampled data for all the actions}
    \label{fig:MSE-Diff}
\end{figure}

\begin{figure}[htbp]
    \centering
    \includegraphics[width=0.75\linewidth]{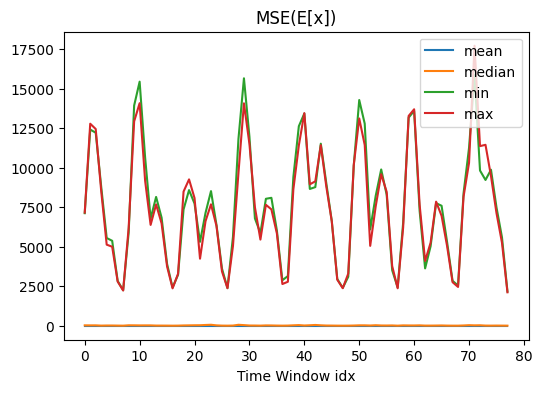}
    \caption{Time windowed analysis for the \textit{Walk} action shownig that also considering small windows the mean downsampling performs better.}
    \label{fig:time_window_analysis_walk}
\end{figure}

Figures \ref{fig:MSE-Diff} and \ref{fig:time_window_analysis_walk} illustrate that the Mean downsampling approach consistently preserved the signal’s core characteristics, even when viewed within smaller time windows or in the frequency domain. Based on these results, the decision was made to proceed with Mean downsampling for the remainder of the work.

\subsection{Modifications to the Architecture}
\label{subsec:architecture}
After preprocessing the \gls{csi} data through downsampling, adjustments to the architecture were required to handle the newly structured data more effectively. The original DeepProbHAR architecture, which processed \gls{csi} data at its full sampling rate, needed to be refined to better accommodate the reduced resolution and to improve classification performance.

The initial architecture (\modeli), used in early iterations, took each 90-frame \gls{csi} segment (which corresponds to a 3-second window) as input. The \gls{csi} data was structured as a matrix \( I_{\rm in} \)
with dimensions \( \textit{frequencies} \times \textit{antennas} \), which was fed into the \gls{lif} spiking neuron layer. The spikes generated by the \gls{lif} layer at each time step were collected and flattened into a one-dimensional array, which was then passed through two fully connected (linear) layers. These layers processed the spiking outputs, and the final classification was performed by a softmax layer.

To improve this basic architecture, two significant modifications were implemented:

\begin{enumerate}
    \item Averaging Over Antennas: Before passing the input to the \gls{lif} layer, the \gls{csi} data was averaged across the antennas using a convolutional layer with a \(1\times1\) kernel. This operation reduced the input dimensionality, collapsing the data from \( \textit{frequencies} \times \textit{antennas} \) to just \( \textit{frequencies} \), making the model simpler and less computationally expensive.

    \item Linear Layer Preprocessing: A linear layer was introduced before the data entered the \gls{lif} layer. This layer processed the averaged \gls{csi} data and helped reduce noise and highlight important features before spike generation. After the linear layer, the processed data was fed into the \gls{lif} layer, and the output spikes were either averaged over the 90-frame window or only the spikes from the final frame were used for classification.
\end{enumerate}
 
Despite these changes, the performance gains were minimal, prompting further architectural modifications to improve the handling of the temporal aspects of the data.

\modelii. \label{model2}
Recognising that human activity can often be captured effectively with fewer frames, therefore the architecture was further refined by introducing temporal windowing within the 90-frame \gls{csi} sequence. Instead of processing all 90 frames as a single sequence, the \gls{csi} data was divided into 10 intervals, each containing 9 frames, representing 0.3 seconds of data. This segmentation aimed to capture short-term movements that could be indicative of more complex activities.

The new architecture processed these 10 intervals in parallel to enhance the model's ability to detect subtle movements over time. The improved architecture operated as follows:

\begin{itemize}
    \item Averaging Across Antennas: Similar to the previous version, the \gls{csi} data was first averaged across the antennas, reducing the \gls{lif} layer's input dimensionality to \( \textit{frequencies} \). 

    \item Interval-Based Processing: The 90-frame sequence was split into 10 intervals, each consisting of 9 frames. Each of these intervals was processed independently by a Leaky Integrate-and-Fire (LIF) layer, capturing small-scale movements and changes in the \gls{csi} signal. The spikes generated by the \gls{lif} layer for each interval were collected, representing the output for that 0.3-second window.

    \item Parallel Processing of Intervals: The output spikes from all 10 intervals were combined to form a tensor with dimensions \( 10\times\textit{frequencies} \), representing the activity across the entire 3-second window. These outputs were passed through a linear layer, which contained \( hidden_dim \) neurons. This layer further compressed and processed the information from all 10 intervals.

    \item Second \gls{lif} Layer for Temporal Integration: The compressed outputs from the linear layer were fed into a second \gls{lif} layer, which processed the data over 10 time steps (one for each interval). The use of a second spiking layer helped integrate the temporal information across the entire 3-second window, capturing longer-term activity patterns.

    \item Final Classification: After processing by the second \gls{lif} layer, the final spikes (a vector of \( \textit{frequencies} \)) were passed through a softmax layer to produce the classification output. This layer provided the activity classification by calculating probabilities for each activity based on the spiking outputs.
\end{itemize}

In this new architecture the averaging was necessary because the \gls{lif} layer does not accept as input tensors of dimension greater than 2 so to add the interval dimension we had to remove the antenna dimension.

\subsection{Modifications to the Symbolic Component}
\label{videoAnalysis}
In addition to changes in the neural network architecture, adjustments were also made to the symbolic component of the neuro-symbolic architecture. Using the anonymised video dataset processed with VideoPose3D as per \cite{cominelliFusion2024}, we tracked 17 key points on the human body to calculate the angles of various limbs, such as the forearms, upper arms, upper legs, and lower legs. 
For each activity recorded in the dataset (e.g., walking, running, clapping), we computed two key metrics based on the motion of these joints over time:

\begin{enumerate}
    \item Range of Motion: For each joint, we calculated the difference between the minimum and maximum positions over a given time window, indicating how much a particular joint moved during the activity.

    \item Mean of Consecutive Differences: This metric was computed by averaging the differences in joint positions between consecutive frames. This provided a measure of the smoothness and variability of movement over time, capturing details about the intensity and frequency of limb movements.
\end{enumerate}

These metrics were then aggregated by body part (e.g., forearms, upper arms, upper legs, lower legs) to form summarised descriptors for arm and leg movements. The idea was to reduce the complexity of the data while preserving the essential characteristics that distinguish one activity from another. For example, walking typically involves alternating movements of the legs and less frequent movements of the arms, while clapping involves repetitive forearm movement with little to no leg involvement.

Once these movement metrics were calculated, we used them to formulate a set of symbolic rules capable of classifying human activities. The goal was to construct a rule-based classifier that could distinguish between different activities based on simple, interpretable concepts.

The rule development process was iterative, relying on an analysis of the key metrics extracted from the video dataset. Initially, it was hypothesised that high-level rules based on broad categories (such as "arm movement" and "leg movement") could distinguish between different activities. However, after further analysis, it became clear that more granular distinctions were needed to achieve accurate classification. Thus, we divided limb movements into finer categories:
\begin{itemize}
    \item Forearm movement: Evaluated using the Mean of Consecutive Differences metric, this category captured the intensity and smoothness of arm movements. High activity in this category was indicative of actions such as clapping or waving.

    \item Upper leg movement: Calculated using the Range of Motion, this feature tracked the overall extent of leg movement, which is useful in distinguishing between activities like running, walking, and jumping.

    \item Lower leg movement: Also measured by the Range of Motion, this captured movements such as the bending of knees during running or jumping, which could distinguish between various dynamic actions.
\end{itemize}

\begin{figure}[htbp]
    \centering
    \includegraphics[width=\linewidth]{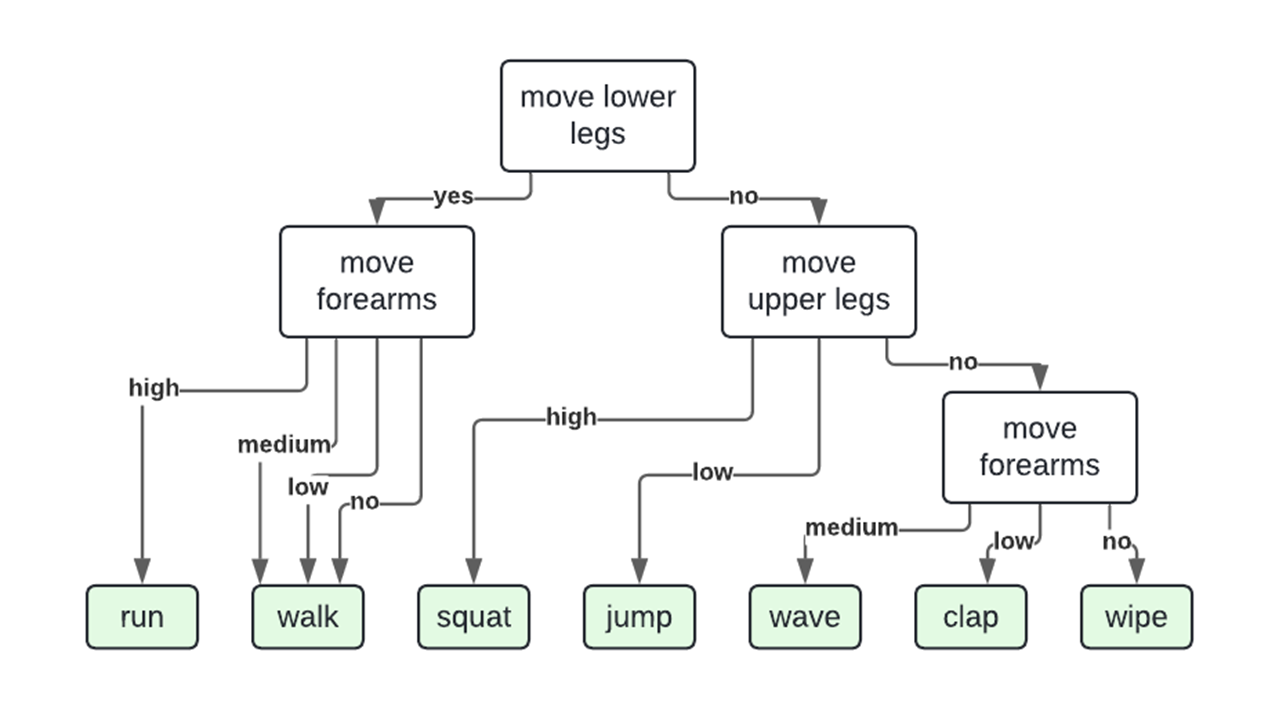}
    \caption{Decision tree form of the rules used to classify activities based on the movement characteristics of different limbs.}
    \label{fig:rule-based-cl}
\end{figure}

Using these metrics, we developed a sequence of rules
that given the quantity of movement of each part say what kind of activity is performed.
Each of these symbolic rules was designed to be simple and interpretable, capturing the essential differences between activities based on a limited set of features.
In Figure \ref{fig:rule-based-cl} we can see these rules in a decision tree form.

After defining the symbolic rules, we integrated them with the neural network outputs using the DeepProbLog framework (see Section \ref{templogic}).

The logic circuit created for DeepProbLog was based on the decision tree of symbolic rules. Each node in the decision tree corresponded to a predicate in the Prolog program, where the predicates represented high-level concepts like ``forearm is moving a lot'' or ``legs are moving.'' The neural network's outputs were treated as probabilistic facts that fed into these predicates. For example, the output of the neural network responsible for detecting forearm movement was used as input to the rule defining clapping, which required high forearm activity and low leg activity.

In this neuro-symbolic system, the neural networks acted as feature detectors, providing the symbolic system with a set of probabilistic inputs about the movement of each body part. The symbolic system then applied logical rules to these inputs to determine the overall activity, combining the flexibility of neural networks with the transparency of symbolic reasoning.

With the symbolic rules established and integrated into DeepProbLog, the next step was refining the neural networks to complement the symbolic reasoning. Instead of using a generic neural network architecture, we modified the \gls{snn} architecture to better align with the symbolic requirements (\modeliii).

In this revised version:
\begin{enumerate}
    \item The spikes generated by the \gls{lif} layers were summed over all time steps rather than just relying on the last frame's spikes. This adjustment ensured that the entire temporal sequence of movements was considered when making predictions about the activity, leading to more robust classification decisions.

    \item The final linear layer from the previous architecture was removed, as it was redundant given the direct input from the summed spikes to the softmax layer. The final classification was performed based on the summed spike outputs, which represented the overall movement activity for a body part. By summing the spikes over the entire sequence, we ensured that even small but consistent movements were captured by the network, improving its sensitivity to subtle activities. This was particularly important for activities like walking, where the movement is continuous but not as pronounced as in activities like jumping or clapping.

    \item The final neuro-symbolic classifier, which utilised \glspl{snn} in combination with symbolic reasoning, was evaluated alongside an alternative neurosymbolic classifier that replaced the \gls{snn} component with a more conventional \gls{cnn}. This comparison allowed for a direct assessment of how the choice of neural architecture impacted the overall performance within the neuro-symbolic framework.
\end{enumerate}

Both the \gls{snn}-based and \gls{cnn}-based classifiers were integrated with the same symbolic reasoning component. The key difference between the two was in the neural part: the \gls{snn} version processed the \gls{csi} data through \gls{lif} layers to generate spike-based outputs, while the \gls{cnn}  version used convolutional layers to extract features from the same data. Despite this architectural difference, both classifiers fed their outputs into the symbolic component, which applied the same set of predefined rules to classify activities.

\section{Temporal Logic}
\label{templogic}
Temporal logic \cite{venema2017temporal} is a formal system designed to reason about time and the sequence of events as they unfold. It is crucial for capturing and representing the dynamic nature of real-world processes, where events occur not only in a specific order but also over defined intervals of time. Temporal logic goes beyond classical propositional logic by introducing time as a dimension for reasoning, which makes it highly suitable for contexts such as human activity recognition where actions take place over distinct periods.

At its core, temporal logic operates on models where time is represented as a series of ordered points or intervals. These can either be discrete (where time progresses in steps) or dense (where there is always another time point between any two others). Time is often modelled using structures such as the natural numbers (N) or real numbers (R), with a precedence relation < that determines whether one time point occurs before another. The use of time-related operators such as \textbf{F} (for ``Future'') and P (for ``Past'') allows temporal logic to express statements about events occurring at different points in time.

For example, \textbf{F} p means ``at some future time, proposition p will be true,'' while P q means ``at some point in the past, proposition q was true.'' These operators enable temporal logic to capture the sequence of events, whether they occur in the past, present, or future. More advanced operators, such as G (for ``Globally'') and H (for ``Historically''), express that a condition holds across all future or past time points, respectively. Thus, temporal logic provides a powerful framework for reasoning about how events unfold over time.

In the context of this work, temporal logic offers a tool for distinguishing between different human actions based on the temporal patterns observed in \gls{csi} data. Human activities, such as walking, running, or jumping, have distinct temporal characteristics that unfold over time, and temporal logic provides a way to formalise and reason about these patterns. For instance, walking might show a periodic pattern in the \gls{csi} data that is completely different from the one shown by wiping. Temporal logic can help define rules that capture these temporal sequences, facilitating the recognition of activities based on how the signal varies over time.

However, while DeepProbLog is a powerful tool for combining probabilistic reasoning with neural networks, it is fundamentally based on Prolog, a system designed to work with a fragment of first-order logic without modalities. To align with this structure and simplify our approach, we decided to use a version of temporal logic that could be more easily translated into first-order logic. This simplified temporal logic would allow us to maintain consistency with the underlying logic framework of Prolog while providing a clear and practical method for reasoning about time-dependent data in a rule-based manner.

Our approach leverages \gls{ltl} \cite{huth2004logic}, a formal system specifically designed for reasoning about sequences of events over time, particularly well-suited for discrete time intervals, as in the case of \gls{csi} data. In this framework, we designed a logic system that operates on these discrete intervals, using conditional rules that reflect the temporal progression of actions. The fundamental structure of our logic follows the \gls{ltl} pattern: \textit{“if during the first part of the interval, condition \textbf{A} holds, and in the second part, condition \textbf{B} holds, then the action is classified as \textbf{Y}”}. This approach mirrors the way \gls{ltl} captures the evolution of states, where temporal operators like G (Globally), \textbf{F} (Future), and \textbf{X} (Next) allow for precise reasoning about sequences of events.

\gls{ltl} operates on propositions that are either true or false at specific time points. In our context, we use \gls{ltl}’s temporal operators to model relationships between different time intervals within the \gls{csi} data. For instance, the \textbf{X} (Next) operator can express that \textit{“if \textbf{A} holds in the current interval, \textbf{B} must hold in the next interval”}. This allows us to capture the sequential nature of human activities, where one event follows another in time.

By integrating the \textbf{X} (Next) operator from \gls{ltl}, we refine the logic by establishing temporal dependencies between actions. For example, \textit{“if condition \textbf{A} holds and condition \textbf{X} (A) holds, then the activity is classified as \textbf{Y}”}. This reflects the core principle of \gls{ltl}, where the \textbf{X} (Next) operator specifies that a certain condition must hold in the immediate next state. In the context of activity recognition, this operator is particularly valuable for modelling how specific movements follow each other in quick succession, as observed in the \gls{csi} data. The \textbf{X} (Next) operator allows us to precisely track the progression of events from one time interval to the next, improving both the accuracy and granularity of the classification process.

One of the key advantages of using \gls{ltl} in this context is its compatibility with the discrete nature of \gls{csi} data, where each sample represents a specific time step. By applying \gls{ltl} operator over these time steps, we can effectively model how different activities manifest through temporal changes in the data. For instance, walking may exhibit regular, predictable transitions over consecutive intervals, while activities like jumping could show more abrupt temporal changes. The \textbf{X} operator allows us to track these patterns and ensure that activities are accurately recognised as they evolve.

Additionally, our approach aligns with the core strengths of \gls{ltl}, which is designed to be easily translatable into first-order logic. This ensures that our system remains compatible with the underlying Prolog framework used in our reasoning engine, which operates on first-order logic. By structuring the temporal relationships in a way that can be directly expressed through \gls{ltl} operators, we provide a transparent and interpretable framework for reasoning about time-dependent data. For example, the “after” operator in \gls{ltl} can be directly mapped to a first-order logic rule such as “if \textbf{X} holds now, then \textbf{Y} must hold in the future,” ensuring a clear and understandable representation of temporal dependencies.

\glsresetall
\section*{Summary}
In this chapter we detail the methodology for enhancing \gls{har} using \wifi \gls{csi} by integrating neural networks and symbolic reasoning. It begins with adjustments to the \gls{csi} data handling, such as downsampling taking the mean value to better align with video frame rates.

The chapter then describes the neural network architecture designed using \gls{snn} and the symbolic component that uses limb movement metrics to generate interpretable rules that classify human activities. These rules are integrated with the neural network output via DeepProbLog, blending neural and symbolic approaches.

Finally, temporal logic is introduced to capture the sequential nature of human actions. This approach allows the system to reason about time-dependent patterns in \gls{csi} data, enhancing the model’s interpretability and accuracy in recognizing activities.

\glsresetall
\chapter{Experimental design and results}
\label{chp:results}
\section*{}
In the preceding chapters, we explored foundational concepts in \gls{har} using \wifi \gls{csi} (Section \ref{dataset}), focusing on different neural network models such as \glspl{cnn} and \glspl{snn} (Section \ref{networks}), and introduced neurosymbolic approaches for enhanced adaptability. In this chapter, we present a detailed experimental evaluation of the proposed methodologies, including a comparison between baseline \gls{cnn} and \gls{snn} models, as well as the statse-of-the-art neurosymbolic classifier. These experiments aim to assess model performance using various metrics, including accuracy, confusion matrices, and Bayesian Hypothesis Testing under multiple scenarios and experimental conditions.
\glsresetall

\section{Baseline SNN vs CNN}
This section presents a comparative analysis of a \gls{snn} \ref{model2} and a simple \gls{cnn} used as benchmark for human activity recognition using \wifi \gls{csi}. Both models were trained and tested under the same experimental conditions, and their performance was evaluated across 10 trials. The analysis covers classification accuracy, confusion matrices, and a statistical comparison using Bayesian Hypothesis Testing \cite{barber:hypothesis:rr:04:57}.

Before the trials, Optuna\footnote{\url{https://optuna.org/} on 27 September 2024.} was used to perform hyperparameter optimisation for both models. Below are the key hyperparameters that were tuned for each model:

SNN Hyperparameters:
\begin{itemize}
    \item \textbf{lr (learning rate)}: The rate at which the model updates its weights during training. 
    \item \textbf{epochs}: The number of complete passes through the training dataset. 
    \item \textbf{step}: The number of time steps passed to the Leaky Integrate-and-Fire (LIF) layer.
    \item \textbf{hidden\_dim}: The number of neurons in the hidden layer. 
    \item \textbf{epoch\_annealing}: controls the increasing importance of the KL divergence in the model's custom loss functions as training progresses. Early in training, the model focuses on minimising the squared loss, while later, the KL divergence helps refine the model’s ability to handle uncertainty in predictions. 
    \item \textbf{reset\_mechanism}: Defines how the neuron membrane potential is reset after firing. Options: 
    \begin{itemize} 
        \item \textbf{zero}: Resets the membrane potential to 0 after firing. 
        \item \textbf{subtract}: Subtracts the threshold from the potential, allowing for a more gradual reset. 
    \end{itemize} 
\end{itemize}

CNN Hyperparameters:
\begin{itemize} 
    \item \textbf{lr (learning rate)}: The rate at which the model updates its weights during training. 
    \item \textbf{epochs}: The number of complete passes through the training dataset. 
    \item \textbf{last\_channel}: Determines the number of channels in the final convolutional layer. It also sets the dimension of the linear layers following the convolutional layers and determines the channels of previous convolutional layers by dividing the value by 1.5 for each preceding layer. 
    \item \textbf{epoch\_annealing}: Similar to the \gls{snn}, controls the importance of the KL divergence in the loss function. 
\end{itemize}

\begin{figure}[htbp]
    \centering
    \begin{subfigure}[b]{0.45\textwidth}
        \includegraphics[width=\textwidth]{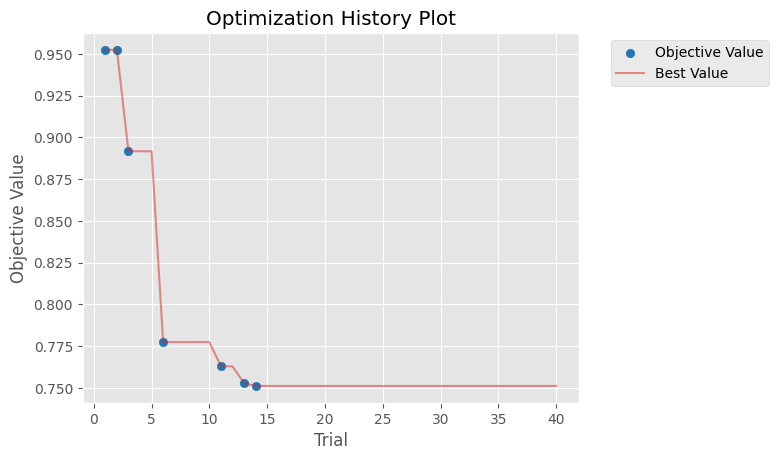}
        \caption{The CNN achived the best loss on trial 14.}
        \label{fig:optunacnn}
    \end{subfigure}
    \begin{subfigure}[b]{0.45\textwidth}
        \includegraphics[width=\textwidth]{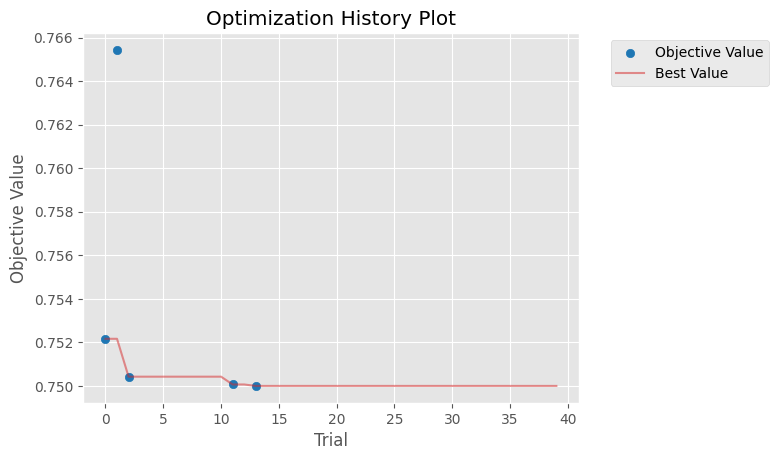}
        \caption{The SNN achived the best loss on trial 13.}
        \label{fig:optunasnn}
    \end{subfigure}
    \caption{Graph that shows the values of the best loss of each optuna trial during the optimisation of both CNN and SNN models.}
    \label{fig:optuna}
\end{figure}

In table \ref{tab:snn_cnn_params} the best hyperparameters for each model obtained after conducting the optimisation (Figure \ref{fig:optuna}).
\begin{table}[htbp]
    \centering
    \begin{tabular}{|l|c|c|}
        \hline
        Parameter           & SNN                          & CNN                          \\ \hline
        \textbf{step}            & 5                            & -                            \\ 
        \textbf{lr}              & 0.006540        & 0.000306        \\ 
        \textbf{epochs}          & 35                           & 35                           \\ 
        \textbf{hidden\_dim}      & 50                           & -                            \\ 
        \textbf{epoch\_annealing} & 30                           & 27                           \\ 
        \textbf{reset\_mechanism} & 'zero'                       & -                            \\ 
        \textbf{last\_channel}    & -                            & 65                           \\ \hline
    \end{tabular}
    \caption{Best hyperparameters for SNN and CNN models}
    \label{tab:snn_cnn_params}
\end{table}

After training both models over 10 trials with the optimised hyperparameters, the \gls{cnn} achieved an average accuracy of \textbf{0.882}, while the \gls{snn} outperformed it with an average accuracy of \textbf{0.906}.

\begin{figure}[htbp] \centering 
    \begin{subfigure}[b]{0.45\textwidth} 
        \centering 
        \includegraphics[width=\textwidth]{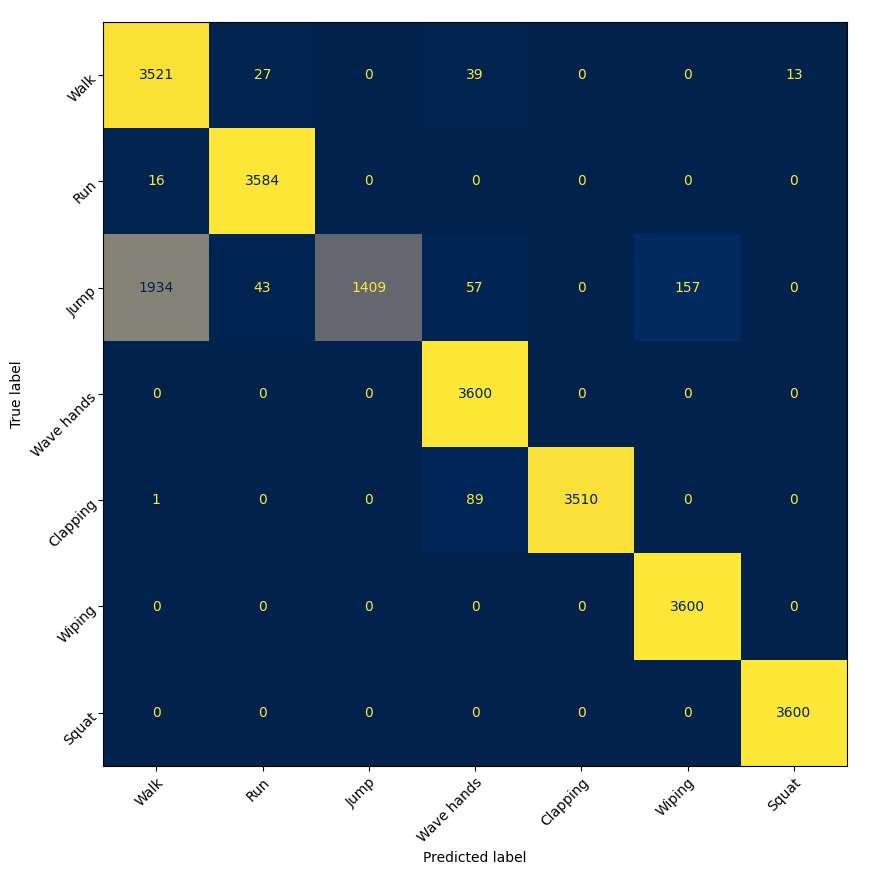} 
        \caption{Confusion matrix for the SNN, showing improved performance with fewer misclassifications, especially in distinguishing between \textit{wave hands} and \textit{clapping} activities.} 
        \label{fig:confMatSnn}
    \end{subfigure} 
    \hfill 
    \begin{subfigure}[b]{0.45\textwidth} 
        \centering 
        \includegraphics[width=\textwidth]{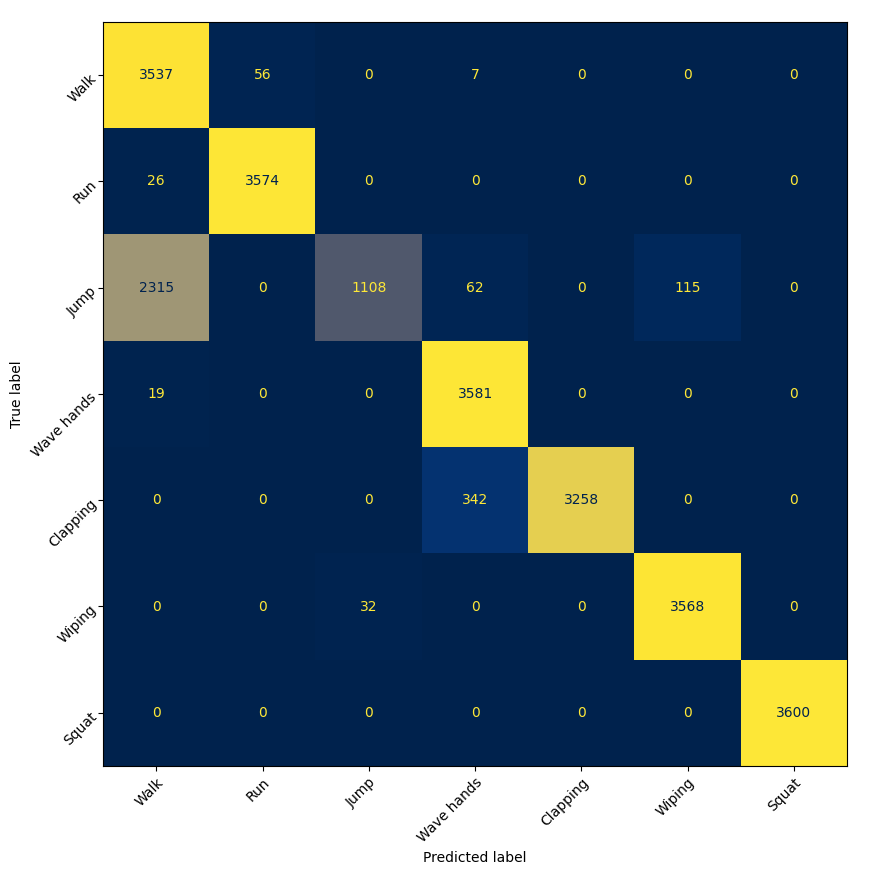} 
        \caption{Confusion matrix for the CNN, illustrating significant misclassifications, particularly between the \textit{wave hands} and \textit{clapping} activities.} 
        \label{fig:confMatCnn} 
    \end{subfigure} 
    \caption{Confusion Matrices for SNN and CNN Classifiers.} 
    \label{fig:CNNvsSNN} 
\end{figure}

The confusion matrices in Figure \ref{fig:CNNvsSNN} illustrate the classification performance across the trials. The \gls{cnn} (\ref{fig:confMatCnn}) struggled particularly with the \textit{Jump} activity, where 2315 instances were misclassified as \textit{Walk}, 115 as \textit{Wiping} and 62 as \textit{Wave hands}. Furthermore, there was some confusion between \textit{Clapping} and \textit{Wave hands}, with 342 \textit{Clapping} activities misclassified as \textit{Wave hands}. The \gls{cnn} performed well in recognising simpler activities like \textit{Run}, \textit{Wave hands}, \textit{Wiping} and \textit{Squat} with minimal errors.

In comparison, the \gls{snn} confusion matrix (\ref{fig:confMatSnn}) demonstrated superior performance, especially in handling the \textit{Jump} activity, though it still misclassified 2191 instances as \textit{Walk}, in most cases, and \textit{Wiping}. There was significantly less confusion between \textit{Clapping} and \textit{Wave hands} activities, with only 89 \textit{Run} activities misclassified as \textit{Walk}. The \gls{snn} performed exceptionally well in classifying activities like \textit{Run}, \textit{Wave hands}, \textit{Clapping}, \textit{Wiping}, and \textit{Squat}, with only a few spares misclassifications observed in those categories.

To statistically validate the performance difference, we employed \textbf{Bayesian Hypothesis Testing} \cite{barber:hypothesis:rr:04:57} (see also Appendix \ref{app:BHT}), a probabilistic method that evaluates the relative likelihood of the two classifiers' performances. This approach is more informative than traditional hypothesis testing, as it provides posterior probabilities, offering a richer understanding of model comparison.

The accuracy of the \gls{snn} was 0.906, which outperformed the \gls{cnn}, which achieved an accuracy of 0.882. Despite the \gls{snn} achieving a higher accuracy, our statistical test suggest that there is overwhelming evidence that the classifiers are indeed different.

The Bayesian analysis emphasizes that these models behave differently when considering more than just their final classification outcomes. The \gls{snn} achieved better accuracy overall, particularly in distinguishing harder-to-classify activities like Wave hands and Clapping, which typically involve more subtle differences in movement patterns.

Initially, this experiment was conducted using independent sequences from the same recording for both the training and testing sets. In this setup, the network learned to distinguish between activities effectively, as reflected in the high accuracy scores and favourable Bayesian analysis.

However, to further test the generalisation capabilities of the models, additional experiments were conducted where the test set consisted of different recordings. These recordings varied in key aspects such as:

\begin{enumerate} \item \textbf{Different subject}: A recording of a different person performing the activities. \item \textbf{Different room}: The activities were performed in a different environment. \item \textbf{Different day}: The same subject and activities, but recorded on a different day. \item \textbf{Same conditions as training but recorded an hour later} \end{enumerate}

As anticipated, the performance of both the \gls{cnn} and \gls{snn} dropped dramatically in these scenarios and thus we omit them here as they are not particularly informative. The network, despite its high performance in distinguishing activities within the same recording, struggled when applied to new recordings, even those with minor variations such as the \textit{same conditions but recorded an hour later}.

This decline in performance indicates that the models were relying heavily on features specific to the recording environment or subject, rather than learning more generalisable patterns. The networks had learned to differentiate between activities based on specific elements of the training data that were not robust enough to transfer to new recordings. This overfitting issue suggests that the networks were likely picking up on subtle, non-relevant variations in the training data (e.g., subject-specific gait or room-specific signal patterns), rather than focusing on the underlying activity-related features in the \gls{csi} data.

\section{Neurosymbolic SNN}
This section presents a comprehensive comparison of two neurosymbolic classifiers: the \neurospykehar  classifier (\modeliii Section \ref{subsec:architecture}) and the DeepProbHAR\cite{cominelliFusion2024} classifier. Both models were tasked with recognising human activities using \wifi \gls{csi} data. We evaluated their performance using several metrics, including accuracy, log likelihood, Bayesian statistics, and confusion matrices to provide an in-depth understanding of each model's strengths and weaknesses.

Other than the use of different neural architectures, a key difference between these two neurosymbolic classifiers lies in their symbolic components, as discussed in Section \ref{videoAnalysis}. Specifically, the \neurospykehar classifier employs a more straightforward symbolic reasoning structure, which is simpler to integrate and manage compared to the more complex architecture of DeepProbHAR. This simpler symbolic reasoning structure permits the \neurospykehar system to rely on only three neural components, whereas DeepProbHAR uses six, reflecting a design philosophy aimed at maintaining simplicity and reducing computational complexity. This simplification also facilitates faster model updates and easier adaptability, which are significant advantages when prioritising efficiency over sophisticated fusion capabilities.

Another notable difference in the methodologies of these two models is related to their training processes. The \neurospykehar model is trained in a single phase, where the data flows directly from input to classification in an end-to-end manner. This approach results in a streamlined training process, which can reduce complexity and computational requirements. On the other hand, the DeepProbHAR classifier uses a two-phase training approach: first, training a \gls{vae} to compress the data into a latent space, and then training a separate classifier to make predictions based on these compressed representations. This two-step process allows the DeepProbHAR model to achieve a better separation of features and more detailed learning from the complex \gls{csi} data.

The accuracy scores for both models reveal their strong capabilities in human activity recognition. The DeepProbHAR classifier achieved an accuracy of 94.29\%, significantly outperforming the \neurospykehar classifier, which achieved 90.45\%. This notable improvement in accuracy can be attributed to the advanced fusion strategies and symbolic integration utilized in DeepProbHAR, which contribute to better differentiation of activities involving nuanced or overlapping movements.

It is also important to note that the \neurospykehar uses a simplified version of the dataset, as discussed in previous sections. Specifically, the \gls{csi} data in \neurospykehar was downsampled, which means that the temporal resolution was reduced to match that of human perceptual capabilities (30 frames per second). Additionally, \neurospykehar employs a simpler management of data from different antennas, resulting in a less comprehensive fusion strategy. In contrast, DeepProbHAR retains the full-resolution \gls{csi} data and implements a more sophisticated fusion process to leverage the data from all antennas effectively.

The Bayesian analysis further supports these findings indicating overwhelming evidence in favor of the DeepProbHAR classifier. This statistical comparison demonstrates that DeepProbHAR has a significantly higher confidence in its predictions and a stronger overall modeling capability compared to \neurospykehar.

\begin{figure}[htbp] \centering 
    \begin{subfigure}[b]{0.45\textwidth} 
        \centering 
        \includegraphics[width=\textwidth]{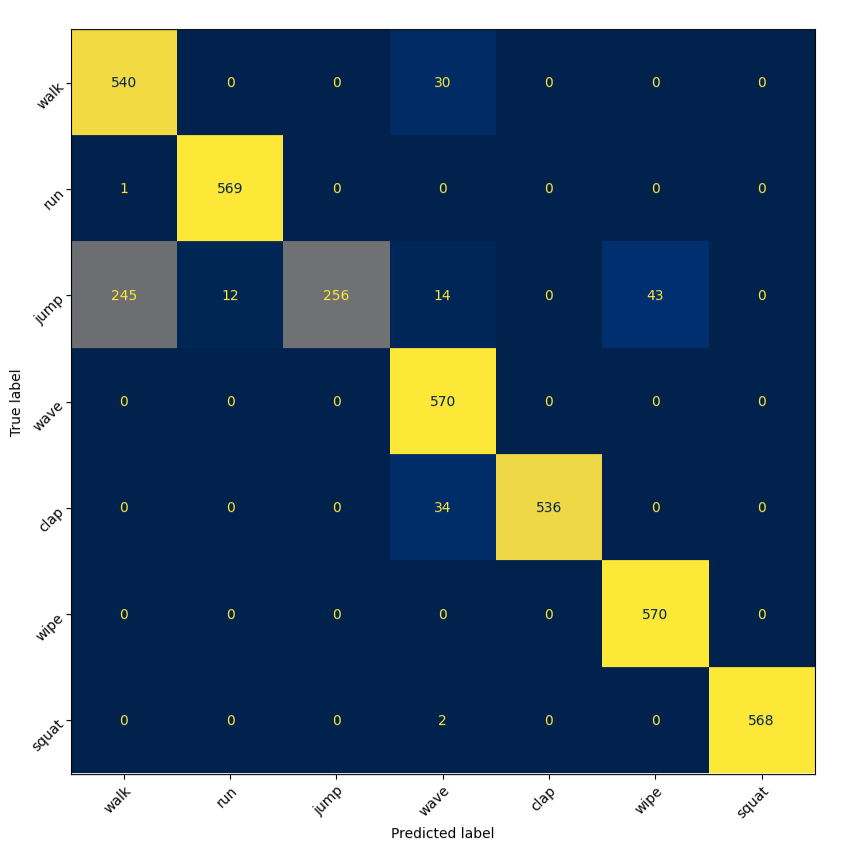} 
        \caption{The confusion matrix for the \neurospykehar classifier shows high accuracy in most activities but includes notable misclassifications, such as \textit{jumping} being confused with \textit{walking}, due to similar movement patterns.} 
        \label{fig:confMatNeuroSpykeHAR} 
    \end{subfigure} 
    \hfill 
    \begin{subfigure}[b]{0.45\textwidth} 
        \centering 
        \includegraphics[width=\textwidth]{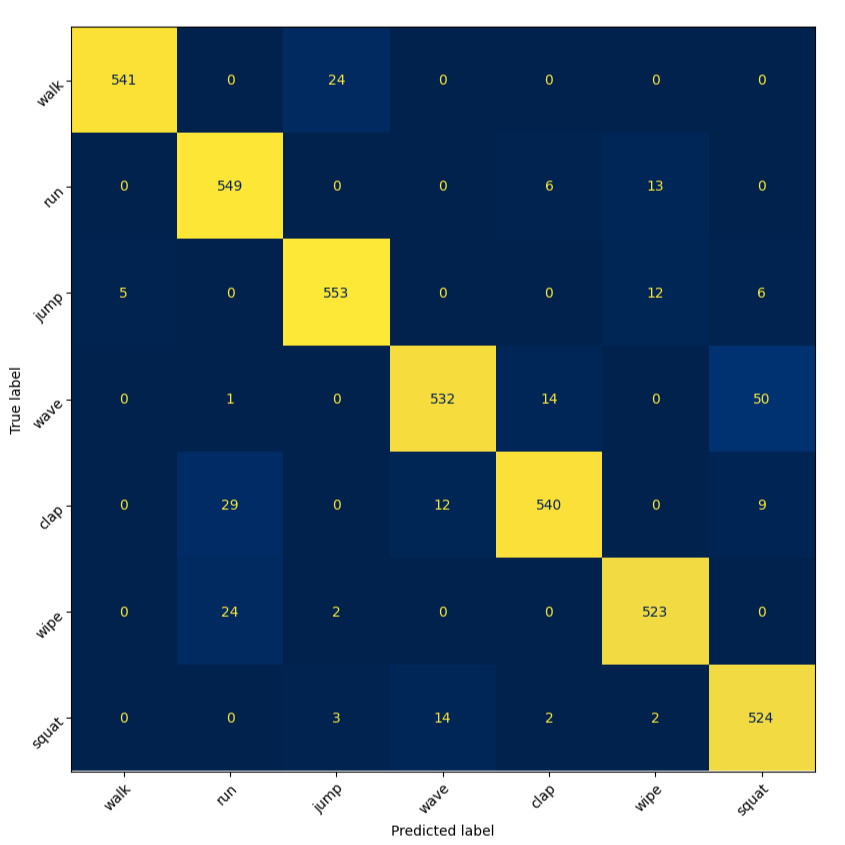} 
        \caption{The confusion matrix for the DeepProbHAR classifier shows fewer misclassifications overall, with better differentiation of activities involving subtle movements, such as \textit{jumping} and \textit{walking}.} 
        \label{fig:confMatDeepProbHAR}
    \end{subfigure} 
    \caption{Confusion Matrices for the \neurospykehar and DeepProbHAR Classifiers
    Confusion matrices comparing the performance of the \neurospykehar and DeepProbHAR classifiers in recognising human activities using \wifi \gls{csi} data.} 
    \label{fig:NSSNNvsDeepProbHAR} 
\end{figure}

The confusion matrices provide a more granular view of each model's performance. For the \neurospykehar classifier (Figure \ref{fig:confMatNeuroSpykeHAR}), the confusion matrix demonstrates its robustness in recognizing the majority of activities accurately, with minimal errors. Activities such as running, waving, clapping and squatting were classified with almost perfect accuracy. However, some confusion is evident, particularly in the jump activity, which was occasionally misclassified as walk. This confusion can be explained by the inherent similarity between the lower body movements of these activities. Despite this, the \neurospykehar classifier still showed consistent results and maintained a relatively low number of misclassifications.

In comparison, the DeepProbHAR classifier (Figure \ref{fig:confMatDeepProbHAR}) also performed well, with even fewer instances of misclassification. It excelled particularly in activities such as jumping, which the \neurospykehar found challenging. For instance, jumping was frequently confused with running in the \neurospykehar model, while DeepProbHAR showed a much clearer separation between these activities. The ability of DeepProbHAR to reduce the rate of misclassification can be attributed to its comprehensive integration of symbolic reasoning and advanced probabilistic modeling, which allows it to effectively leverage the temporal and spatial relationships present in the data.

DeepProbHAR's higher accuracy and fewer misclassifications highlight its effectiveness in distinguishing complex human activities with overlapping features. The \neurospykehar, while highly effective and competitive, occasionally struggled with distinguishing between activities with similar temporal patterns, such as jumping versus walking. However, it still provided a strong performance, reflecting the strength of Spiking Neural Networks in capturing temporal dynamics.

It is worth noting that the \neurospykehar classifier utilized a simplified version of the dataset, which involved downsampling the data and employing a simpler approach to managing the multiple antenna signals, and a reduced number of neural networks. These simplifications could potentially impact its ability to distinguish between more intricate activities, as it results in reduced temporal resolution and less detailed spatial information. Additionally, the training process for \neurospykehar is carried out in a single phase, where the model learns directly from input data to classification, making it a more straightforward, albeit slightly less nuanced, approach compared to DeepProbHAR. DeepProbHAR employs a two-phase training process, where the VAE is first trained to encode the data into a compressed latent space, followed by the training of a separate classifier, which enables more refined learning and better performance.

\section{Temporal Neurosymbolic}
In this section, we explore the causal relationships between movement patterns across various body segments during human activities using the \gls{lpcmci} algorithm \cite{NEURIPS2020_94e70705}, implemented with the \gls{cmis} test \cite{pmlr-v84-runge18a} as the conditional independence test (further details on Section \ref{subsec:casualdisc}). This causal analysis was conducted on the video data discussed in Section \ref{videoAnalysis}, which captures detailed motion patterns across multiple body segments during different activities.

The input data is organised into a set of binary time-series variables, each representing whether a specific body segment was in motion during each time step. The variables are:

\begin{itemize}
    \item \textbf{move upperarms (True/False)}: Indicates whether the upper arms were moving.
    \item \textbf{move forearms a lot (True/False)}: Indicates whether the forearms were moving with significant amplitude.
    \item \textbf{move forearms (True/False)}: Indicates whether the forearms were moving.
    \item \textbf{move upperleg a lot (True/False)}: Indicates whether the upper legs were moving with significant amplitude.
    \item \textbf{move upperleg (True/False)}: Indicates whether the upper legs were moving.
    \item \textbf{move lowerleg (True/False)}: Indicates whether the lower legs were moving.
\end{itemize}

These binary variables were extracted from video data by tracking the movements of different body segments. Similar to the process discussed in Section \ref{videoAnalysis}, this involved analysing the video to classify whether each body part was moving or moving a lot at each time point. This form of binary data was then used as input for the causal discovery analysis.

\subsection{Causal Discovery in Time Series}
\label{subsec:casualdisc}
\textbf{Tigramite}\footnote{\url{https://github.com/jakobrunge/tigramite} on September 2024.} \cite{runge2023causal} is an open-source Python package designed for causal discovery in multivariate time series. It is particularly useful for inferring both direct and indirect causal relationships in systems where dependencies may be nonlinear, time-lagged, and influenced by hidden variables. The Tigramite framework allows for the detection of causal networks by testing conditional independencies among variables, which are the key to identifying true causal structures in complex systems.

The primary causal discovery algorithm used in this analysis is \gls{pcmci} \cite{doi:10.1126/sciadv.aau4996}, and its extension \gls{lpcmci} \cite{NEURIPS2020_94e70705}, which accounts for hidden confounders. \gls{pcmci} detects time-lagged dependencies between variables and estimates causal networks by iteratively testing for conditional independence across various time lags. This method ensures that the detected causal links are not simply spurious correlations but are statistically significant, given the available information and potential confounders.

PCMCI performs the following steps:
\begin{enumerate}
    \item Parent and Children (PC) selection: For each variable, \gls{pcmci} identifies a set of potential causes by performing conditional independence tests over lagged variables. This step selects a pool of variables (lagged and contemporaneous) that are most likely to influence the target variable.
   
    \item Momentary Conditional Independence (MCI) tests: After selecting the PC set, \gls{pcmci} applies MCI tests to examine whether the relationship between two variables, conditioned on their parent variables, holds significant. This step refines the causal graph by confirming or rejecting potential causal links based on these conditional tests.
\end{enumerate}

LPCMCI extends \gls{pcmci} by explicitly handling the presence of latent (hidden) confounders. This is particularly relevant in the video data used in this analysis, where some factors that influence movement patterns might not be directly observable. \gls{lpcmci} adjusts for these unobserved influences.

To perform the independence tests required by PCMCI, we used the \gls{cmis} test \cite{pmlr-v84-runge18a}, which is based on the information-theoretic concept of Conditional Mutual Information (CMI). CMI measures the dependency between two variables, conditioned on a third variable (or set of variables), without assuming any specific parametric form for the dependencies. This makes it particularly suited for handling nonlinear relationships between variables, which are common in human movement data.

CMI is computed as:

\[
I(X;Y|Z) = \sum p(z) \sum \sum p(x,y|z) \log \frac{p(x,y|z)}{p(x|z) \cdot p(y|z)} \, dx \, dy \, dz
\]

\noindent where:
\begin{itemize}
    \item \(X\) and \(Y\) are the two variables being tested for dependency.
    \item \(Z\) is the conditioning variable or set.
    \item \(p(x, y | z)\) is the joint probability of \(X\) and \(Y\), conditioned on \(Z\).
    \item \(p(x | z)\) and \(p(y | z)\) are the marginal probabilities of \(X\) and \(Y\), respectively, conditioned on \(Z\).
\end{itemize}

The local shuffle test used in \gls{cmis} generates a null distribution under the assumption of independence, helping to control for potential false positives. This method ensures that only statistically significant dependencies are identified as causal links, making the results more reliable.

\subsection{Interpretation of Causal Graphs and Results}

LPCMCI, combined with the CMISymb test, produces a causal graph representing the relationships between body segment movements over time, as captured in the binary time-series data extracted from the video recordings. Each node in the graph corresponds to a body segment (e.g., "move upperarms" or "move forearms a lot"), and each directed edge indicates a causal influence from one body segment to another.

The edges in the graph represent causal relationships, which can either be:
\begin{itemize}
    \item Contemporaneous (t = 0): A body segment's movement directly causes the movement of another body part within the same time step.
    \item Lagged (t - n): A body segment’s movement at a previous time step influences another body segment's movement at a future time step.
\end{itemize}
In this work we just considered lagged relationships with n from 1 to 3 as we were looking for relations that could have been easily translated to a temporal logic to use as explain in Section \ref{templogic}.

Additionally, the strength of the causal relationships is represented by the MCI values, which are derived from the conditional independence tests. These values quantify the degree of dependency between the body segments, after accounting for possible confounding factors.

\begin{figure}[htbp]
    \centering
    \begin{subfigure}[b]{0.45\textwidth}
        \centering
        \includegraphics[width=\textwidth]{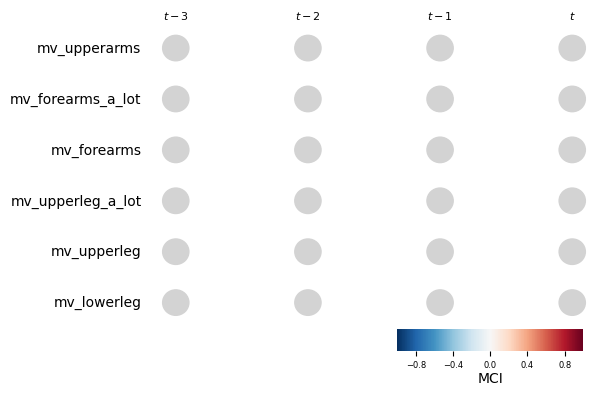}
        \caption{Causal graph of \textit{clap} activity}
        \label{fig:causalClap}
    \end{subfigure}
    \hfill
    \begin{subfigure}[b]{0.45\textwidth}
        \centering
        \includegraphics[width=\textwidth]{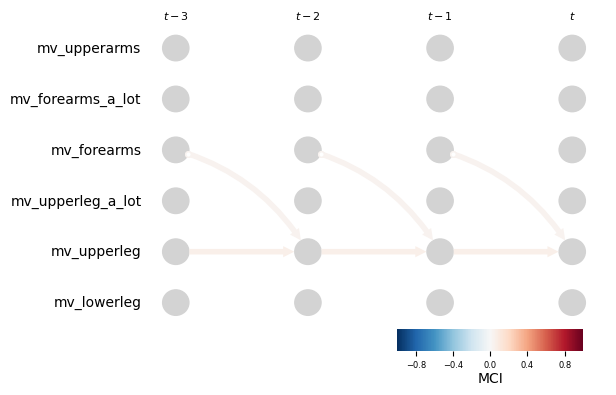}
        \caption{Causal graph of \textit{jump} activity}
        \label{fig:causalJump}
    \end{subfigure}
    \hfill
    \begin{subfigure}[b]{0.45\textwidth}
        \centering
        \includegraphics[width=\textwidth]{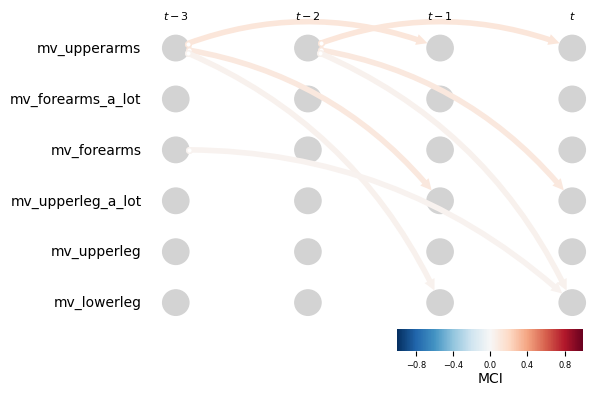}
        \caption{Causal graph of \textit{run} activity}
        \label{fig:causalRun}
    \end{subfigure}
    \hfill
    \begin{subfigure}[b]{0.45\textwidth}
        \centering
        \includegraphics[width=\textwidth]{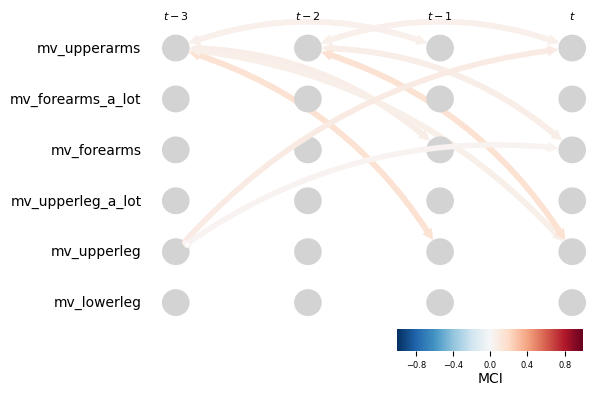}
        \caption{Causal graph of \textit{squat} activity}
        \label{fig:causalSquat}
    \end{subfigure}
    \hfill
    \begin{subfigure}[b]{0.45\textwidth}
        \centering
        \includegraphics[width=\textwidth]{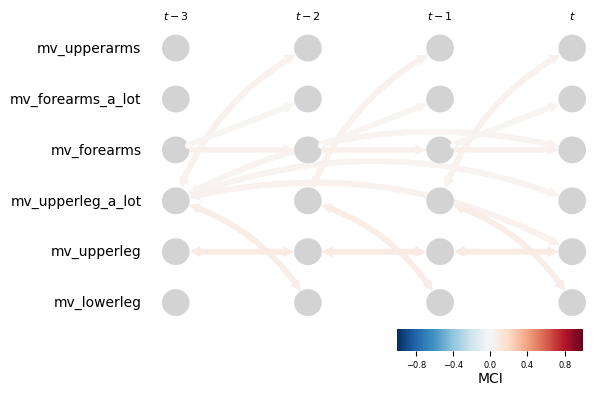}
        \caption{Causal graph of \textit{walk} activity}
        \label{fig:causalWalk}
    \end{subfigure}
    \hfill
    \begin{subfigure}[b]{0.45\textwidth}
        \centering
        \includegraphics[width=\textwidth]{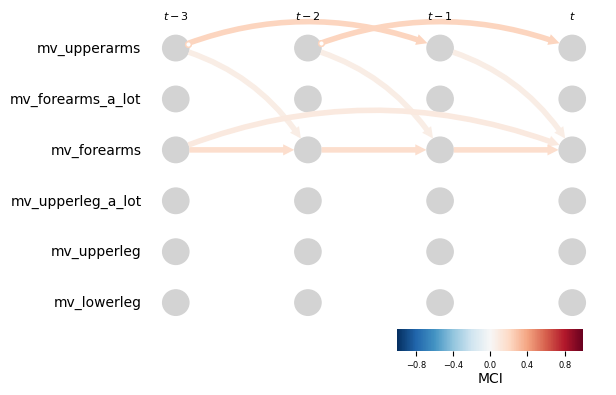}
        \caption{Causal graph of \textit{wave} activity}
        \label{fig:causalWave}
    \end{subfigure}
    \hfill
    \begin{subfigure}[b]{0.45\textwidth}
        \centering
        \includegraphics[width=\textwidth]{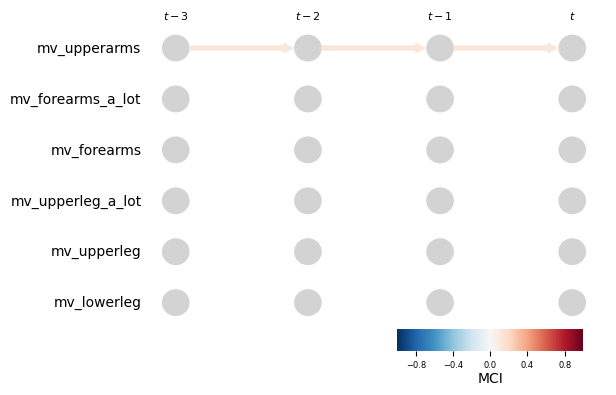}
        \caption{Causal graph of \textit{wipe} activity}
        \label{fig:causalWipe}
    \end{subfigure}
    \caption{Causal graphs generated using LPCMCI with CMISymb}
\end{figure}

The causal graphs generated using \gls{lpcmci} with \gls{cmis} provide insights into the relationships between different body segments during various activities, as recorded in the video data. Here, we highlight some key findings for different activities:
\begin{enumerate}
    \item Clapping (Fig. \ref{fig:causalClap}): The causal graph for clapping shows strong directed edges from the upper arms to the forearms, consistent with the movement pattern in this action. The minimal involvement of the legs is expected, as clapping is primarily an upper-body movement.

    \item Jumping (Fig. \ref{fig:causalJump}): Jumping reveals strong bidirectional relationships between the upper legs and lower legs. These causal links show the necessary coordination between the legs during the propulsion and landing phases of the jump.

    \item Running (Fig. \ref{fig:causalRun}): Running demonstrates a more complex causal network involving both the arms and legs. The feedback loops between the upper arms and forearms, as well as between the upper legs and lower legs, suggest that running is a highly coordinated full-body movement.

    \item Squatting (Fig. \ref{fig:causalSquat}): Squatting, which involves primarily the legs, shows strong causal relationships between the lower legs and upper legs, with minimal upper-body involvement. This aligns with the biomechanics of squatting, where balance and force generation are driven by the lower body.
    
    \item Walking (Fig. \ref{fig:causalWalk}): Walking exhibits balanced causal links between the arms and legs. The arrows from the upper arms to the forearms show the importance of arm movement for balance, while the causal links between the upper legs and lower legs reflect the coordinated leg movement needed for locomotion.

    \item Waving (Fig. \ref{fig:causalWave}): Waving, being an upper-body-focused activity, shows strong causal relationships between the upper arms and forearms, with minimal leg involvement. This is expected, as waving involves repetitive arm movements without much lower body motion.

    \item Wiping (Fig. \ref{fig:causalWipe}): Similar to waving, wiping shows strong causal links between the upper arms and forearms, reflecting the coordinated, repetitive upper-body motion required for the task.
\end{enumerate}

This causal analysis provides a comprehensive view of the temporal and directional dependencies between different body segments during various activities. The use of \gls{lpcmci}, with its ability to handle latent confounders and time-lagged relationships, allowed us to uncover both direct and indirect causal links in this complex system of human movement.

The causal graphs not only highlight the key movement patterns but also reveal feedback mechanisms between different body segments, suggesting that certain movements influence subsequent actions. 

\subsection{Temporal Neurosymbolic results}
We decided to proof-check our approach with a simpler setup by focusing on binary classification between two specific activities: Walk and Squat. This decision allowed us to validate the feasibility and effectiveness of our methodology in a controlled and straightforward scenario. Given the encouraging results from this causal analysis, we aimed to translate the relationships captured by our model into a temporal logic representation to train a neurosymbolic classifier.

We concentrated on Walk and Squat when formulating the temporal logic rules because they presented an interesting challenge, as their underlying graphical representations contained common nodes but differed significantly in their edge structures. This characteristic made them ideal for evaluating the ability of our temporal logic to distinguish between subtle variations in the sequence and transitions of activities, thus providing a solid foundation for testing the neurosymbolic classifier in identifying nuanced differences.

The neurosymbolic classifiers, as better explained in section \ref{templogic}, incorporate two key components:
\begin{itemize}
    \item Neural Networks (SNN/CNN): Responsible for perceiving patterns in the \gls{csi} data.

    \item Logic: This component applies rules defined in a Prolog file to interpret the outputs of the neural networks. 
\end{itemize}   

For this classifier the temporal logic reasoning is based on simple rules extracted from figures \ref{fig:causalWalk} and \ref{fig:causalSquat} and are:
\begin{verbatim}
activity(X,walk) :- move_upper_legs_1(X,yes), 
                    move_upper_legs_2(X,yes), 
                    move_upper_legs_3(X,yes), 
                    move_upper_arms_1(X,no).
                    
activity(X,squat)  :- move_upper_arms_1(X,yes), 
                      move_upper_legs_2(X,no), 
                      move_upper_legs_3(X,yes), 
                      move_upper_legs_1(X,no).
\end{verbatim}
\label{temporalLogic}

This hybrid approach allows for precise classification, as the neural network detects patterns and passes probabilistic predictions to the logic rules, which then classify the activities based on predefined temporal constraints.

\begin{figure}[htbp]
    \centering
    \begin{subfigure}[b]{0.45\textwidth}
        \centering
        \includegraphics[width=\textwidth]{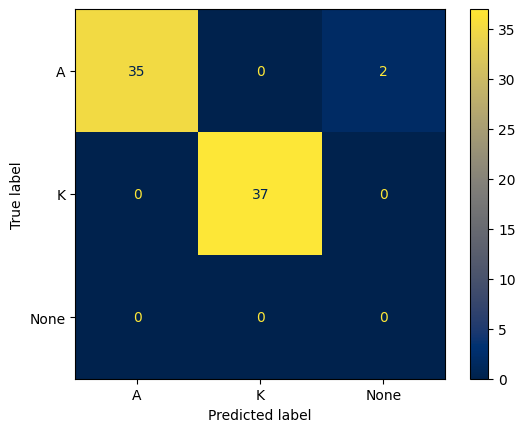}
        \caption{Confusion matrix for the \templogcnn classifier.}
        \label{fig:confMatCNNtemp}
    \end{subfigure}
    \hfill
    \begin{subfigure}[b]{0.45\textwidth}
        \centering
        \includegraphics[width=\textwidth]{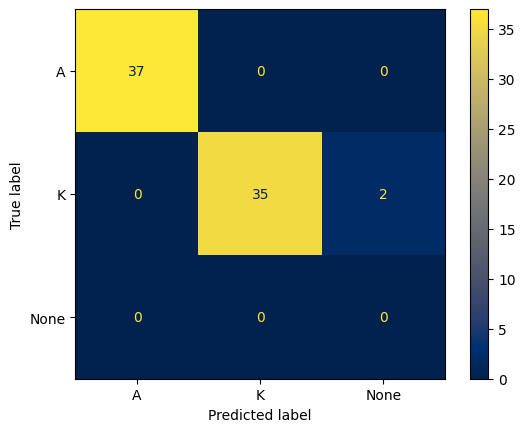}
        \caption{Confusion matrix for the \templogspyke classifier.}
        \label{fig:confMatSNNtemp}
    \end{subfigure}
    \caption{Confusion Matrices for \templogcnn and \templogspyke for walk and squat activities.}
    \label{fig:confMatTemp}
\end{figure}

\begin{figure}[htbp]
    \centering
    \begin{subfigure}[b]{0.45\textwidth}
        \centering
        \includegraphics[width=\textwidth]{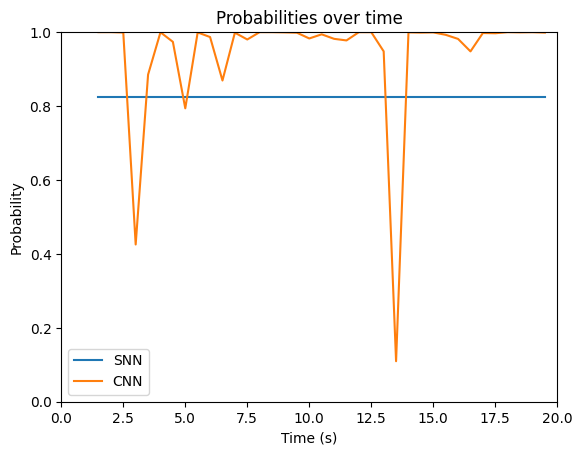}
        \caption{Probabilities of the predictions on the walk activity.}
    \end{subfigure}
    \hfill
    \begin{subfigure}[b]{0.45\textwidth}
        \centering
        \includegraphics[width=\textwidth]{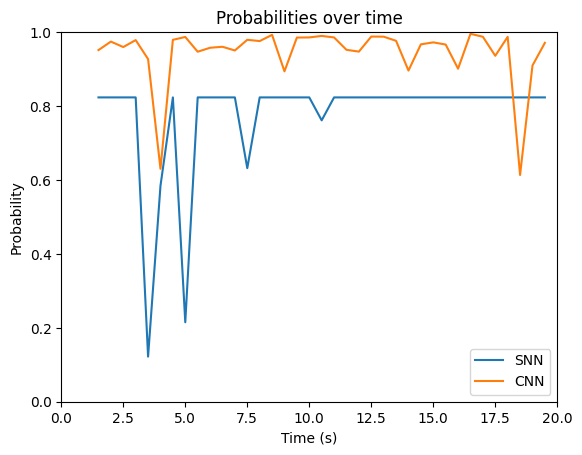}
        \caption{Probabilities of the predictions on the squat activity.}
    \end{subfigure}
    \caption{Probabilities of the predictions of the two classifiers.}
    \label{fig:probabilities}
\end{figure}

The results of these classifiers are summarised in the following confusion matrices (figure \ref{fig:confMatTemp}), providing a comparison between the neurosymbolic classifiers using the \gls{snn} (\templogspyke model) and \gls{cnn} (\templogcnn model) components.
These matrices highlight a similar ability on recognizing the two activities both failing on just 2 cases not predicting the wrong activity but producing as output a low probability for both with the highest being the correct one. 
This is can be seen in figure \ref{fig:probabilities} that shows the prediction probabilities for walking and squatting activities as functions of time, with one plot representing the predictions on the walk activity and the other representing the squat activity.
The fact that the probabilities of the 2 classes doesn't sum to 1 is due to the fact the logic \ref{temporalLogic} doesn't take into account all the possible combinations of \( \textit{move\_apper\_leg\_1} \), \( \textit{move\_apper\_leg\_2} \), \( \textit{move\_apper\_leg\_3} \) and \( \textit{move\_apper\_arms\_1} \).

The results highlight that there might be advantages using neurosymbolic classifiers that incorporate temporal logic for human activity recognition.
By integrating temporal logic into the classification process, the neurosymbolic approach allows for more interpretable and reliable predictions. The temporal rules applied in Prolog ensure that the classifiers not only make decisions based on observed patterns but also adhere to logical constraints related to how these activities unfold over time.

\section*{Summary}
This chapter's evaluation demonstrated the strengths and weaknesses of different neural and neurosymbolic models for activity recognition using \wifi \gls{csi}. While traditional \gls{cnn} and \gls{snn} models each showed unique advantages, the neurosymbolic approach demonstrated the potential for improved interpretability and handling of temporal dependencies. These results underline the necessity of combining neural learning with temporal reasoning for achieving more robust and adaptable \gls{har}.

\glsresetall
\chapter{Conclusions}
\label{chp:conclusions}
\section*{}
In this thesis, we explored innovative approaches to improve the efficiency, adaptability, and interpretability of \gls{har} systems using \wifi \gls{csi}. Our primary objective was to develop advanced models capable of processing \gls{csi} data in a non-intrusive manner, while achieving high recognition accuracy with energy-efficient computation. By employing \glspl{snn} and integrating neurosymbolic reasoning, we addressed the challenges associated with energy consumption and adaptability which are essential for practical \gls{har} applications.

One of the most significant contributions of this research was the successful implementation of \glspl{snn} for \gls{har}. Unlike traditional deep learning models, such as \glspl{cnn}, which operate with continuous activations, \glspl{snn} are inspired by the biological function of neurons, using discrete spikes to represent and transmit information. This fundamental difference enables \glspl{snn} to perform computations in an event-driven manner, thereby enhancing real-time responsiveness and reducing energy consumption. The experimental results clearly demonstrated that the \gls{snn} achieved a competitive average activity recognition accuracy of 90.6\%, compared to 88.2\% for the \gls{cnn}. These results indicate that \glspl{snn} are well-suited for \gls{har}, particularly in resource-constrained environments, such as embedded or edge computing platforms where power efficiency is crucial.

Beyond energy-efficient computation, this thesis also introduced the first neurosymbolic Spiking Neural Network (\neurospykehar) for activity recognition. The integration of symbolic logic with \glspl{snn} adds an essential layer of interpretability, allowing for a clearer understanding of the decision-making process—a feature that is often lacking in purely neural-based models. The \neurospykehar achieved an accuracy of 90.45\%, which is almost comparable to the state-of-the-art neurosymbolic model, DeepProbHAR\cite{cominelliFusion2024}, which achieved 94.29\%. Although the accuracy of the \neurospykehar is slightly lower, its ability to provide explanations for its outputs and adapt to new activities more flexibly represents a meaningful advancement. The neurosymbolic integration allows for efficient updates through the introduction of new logical rules, minimizing the need for costly retraining.

The methodological contributions, as outlined in Chapter \ref{chp:methodology}, include the implementation of temporal logic to capture the sequence of human activities, thus improving the model's temporal understanding. By considering how activities unfold over time, the system was better able to distinguish between similar activities, such as walking versus running, by accurately modeling the temporal relationships between different actions. This improvement in temporal understanding enhanced the robustness of the model, making it suitable for complex \gls{har} scenarios. Chapter \ref{chp:results} presented the results of this work, highlighting the effectiveness of the neurosymbolic approach in providing both efficient computation and interpretability. The integration of neurosymbolic reasoning proved beneficial for not only maintaining recognition accuracy but also for enhancing transparency, which is critical in applications like healthcare and smart environments where understanding model decisions is crucial.

However, this work is not without its limitations. One key limitation lies in the use of a fixed time step for processing spiking events. While using a fixed time step simplifies the computational model, it can reduce the model's ability to fully capture the rich, temporal dynamics inherent in human activities, potentially affecting the resolution and accuracy of activity detection. To address this limitation, future work could involve incorporating variable or adaptive time-step mechanisms that adjust dynamically based on the detected activity. Such an approach would allow the model to adapt its temporal resolution depending on the complexity and speed of the activity, thereby improving accuracy without incurring a significant computational burden. By making the system more responsive to changes in temporal dynamics, we can better capture subtle variations in human behavior, which is particularly important for activities that involve sudden or unpredictable movements.

Another important aspect for future research involves the implementation of the developed models on spiking neuromorphic hardware. This would reveal potential hardware-specific challenges and allow for optimization of the model for real-world deployments. Testing the \neurospykehar on neuromorphic hardware would also open opportunities to further evaluate the integration of complex symbolic logic in a physical system, ultimately leading to practical, deployable \gls{har} solutions.

In addition to deploying the models on neuromorphic hardware, further research should focus on enhancing the neurosymbolic reasoning component. In particular, incorporating more sophisticated temporal logic could further improve the system’s ability to handle intricate sequences of human actions. The current implementation of temporal logic has shown promising results in capturing basic sequential dependencies; however, human activity often involves overlapping or concurrent movements that may require more advanced representations.

Lastly, expanding the dataset and testing in diverse environments is another key future direction. The current research was conducted using a dataset that covered a range of human activities in a controlled setting. However, \gls{har} systems intended for real-world deployment must be robust to variability in conditions such as changes in the environment, the presence of multiple people, and unpredictable activity patterns.

\glsresetall
In conclusion, this thesis has advanced the capabilities of \gls{har} systems by demonstrating the effectiveness of \glspl{snn} for efficient, real-time computation, and introducing the first neurosymbolic \gls{snn} that adds significant interpretability and flexibility. By focusing on the computational efficiency of \glspl{snn} and the explanatory power of symbolic reasoning, this work lays a strong foundation for the future development of practical, adaptable, and transparent \gls{har} systems. As spiking neuromorphic hardware becomes more readily available, the potential to combine low power consumption with rich interpretability will become increasingly achievable, marking an important step toward integrating intelligent \gls{har} systems seamlessly into daily life. The hybrid model presented here enhances the ability of machines to perform complex perceptual tasks while making these processes more traceable, thereby paving the way for more adaptable intelligent systems.

\chapter*{Acknowledgements}
The author thanks Prof. Mani B. Srivastava (ECE Department, University of California), Dr. Lance M. Kaplan (DEVCOM Army Research Lab), Dr. Trevor Bihl, Dr. Erik P. Blasch, and Dr. Nandini Iyer (Air Force Research Laboratory) for their valuable assistance and insights during the development of this work. While preparing this work, the author utilized GPT-4o to enhance readability and language. After using these tools, the author carefully reviewed and edited the content, and he takes full responsibility for the final version of the publication.


\begin{thebibliography}{10}

\bibitem{atzmueller2018explicative}
Martin Atzmueller, Naveed Hayat, Matthias Trojahn, and Dennis Kroll.
\newblock Explicative human activity recognition using adaptive association rule-based classification.
\newblock In {\em 2018 IEEE International Conference on Future IoT Technologies (Future IoT)}, pages 1--6. IEEE, 2018.

\bibitem{bahadori2022rewis}
N.~Bahadori, J.~Ashdown, and F.~Restuccia.
\newblock {{ReWiS}}: Reliable {{Wi-Fi}} sensing through few-shot multi-antenna multi-receiver {{CSI}} learning.
\newblock In {\em Proceedings of IEEE International Symposium on a World of Wireless, Mobile and Multimedia Networks (WoWMoM)}, pages 50--59, 2022.

\bibitem{barber:hypothesis:rr:04:57}
David Barber.
\newblock Are two classifiers performing equally? a treatment using bayesian hypothesis testing.
\newblock Idiap-RR Idiap-RR-57-2004, IDIAP, Rue de Simplon 4, Martigny, CH-1920, Switerland, 5 2004.
\newblock IDIAP-RR 04-57.

\bibitem{boureau2010theoretical}
Y-Lan Boureau, Jean Ponce, and Yann LeCun.
\newblock A theoretical analysis of feature pooling in visual recognition.
\newblock In {\em Proceedings of the 27th international conference on machine learning (ICML-10)}, pages 111--118, 2010.

\bibitem{bridle1990probabilistic}
John~S Bridle.
\newblock Probabilistic interpretation of feedforward classification network outputs, with relationships to statistical pattern recognition.
\newblock In {\em Neurocomputing: Algorithms, architectures and applications}, pages 227--236. Springer, 1990.

\bibitem{chen2012sensor}
Liming Chen, Jesse Hoey, Chris~D Nugent, Diane~J Cook, and Zhiwen Yu.
\newblock Sensor-based activity recognition.
\newblock {\em IEEE Transactions on Systems, Man, and Cybernetics, Part C (Applications and Reviews)}, 42(6):790--808, 2012.

\bibitem{cominelliFusion2024}
Marco Cominelli, Francesco Gringoli, Lance~M. Kaplan, Mani~B. Srivastava, Trevor Bihl, Erik~P. Blasch, Nandini Iyer, and Federico Cerutti.
\newblock Neuro-symbolic fusion of wi-fi sensing data for passive radar with inter-modal knowledge transfer, 2024.

\bibitem{fusion2023}
Marco Cominelli, Francesco Gringoli, Lance~M. Kaplan, Mani~B. Srivastava, and Federico Cerutti.
\newblock Accurate passive radar via an uncertainty-aware fusion of {{Wi-Fi}} sensing data.
\newblock In {\em 26th International Conference on Information Fusion (FUSION)}, 2023.

\bibitem{9927729}
Manon Dampfhoffer, Thomas Mesquida, Alexandre Valentian, and Lorena Anghel.
\newblock Are snns really more energy-efficient than anns? an in-depth hardware-aware study.
\newblock {\em IEEE Transactions on Emerging Topics in Computational Intelligence}, 7(3):731--741, 2023.

\bibitem{davies2018loihi}
Mike Davies, Narayan Srinivasa, Tsung-Han Lin, Gautham Chinya, Yongqiang Cao, Sri~Harsha Choday, Georgios Dimou, Prasad Joshi, Nabil Imam, Shweta Jain, et~al.
\newblock Loihi: A neuromorphic manycore processor with on-chip learning.
\newblock {\em Ieee Micro}, 38:82--99, 2018.

\bibitem{d2020neurosymbolic}
Artur d'Avila Garcez and Luis~C Lamb.
\newblock Neurosymbolic ai: The 3rd wave.
\newblock {\em arXiv e-prints}, pages arXiv--2012, 2020.

\bibitem{de2007problog}
Luc De~Raedt, Angelika Kimmig, and Hannu Toivonen.
\newblock Problog: A probabilistic prolog and its application in link discovery.
\newblock In {\em IJCAI}, volume~7, pages 2462--2467. Hyderabad, 2007.

\bibitem{eshraghian2021training}
Jason~K. Eshraghian, Max Ward, Emre~O. Neftci, Xinxin Wang, Gregor Lenz, Girish Dwivedi, Mohammed Bennamoun, Doo~Seok Jeong, and Wei~D. Lu.
\newblock Training spiking neural networks using lessons from deep learning.
\newblock {\em Proceedings of the IEEE}, 111(9):1016--1054, 2023.

\bibitem{FIERENS_VAN}
DAAN FIERENS, GUY VAN DEN~BROECK, JORIS RENKENS, DIMITAR SHTERIONOV, BERND GUTMANN, INGO THON, GERDA JANSSENS, and LUC DE~RAEDT.
\newblock Inference and learning in probabilistic logic programs using weighted boolean formulas.
\newblock {\em Theory and Practice of Logic Programming}, 15(3):358–401, 2015.

\bibitem{9083980}
Biying Fu, Naser Damer, Florian Kirchbuchner, and Arjan Kuijper.
\newblock Sensing technology for human activity recognition: A comprehensive survey.
\newblock {\em IEEE Access}, 8:83791--83820, 2020.

\bibitem{NEURIPS2020_94e70705}
Andreas Gerhardus and Jakob Runge.
\newblock High-recall causal discovery for autocorrelated time series with latent confounders.
\newblock In H.~Larochelle, M.~Ranzato, R.~Hadsell, M.F. Balcan, and H.~Lin, editors, {\em Advances in Neural Information Processing Systems}, volume~33, pages 12615--12625. Curran Associates, Inc., 2020.

\bibitem{pmlr-v15-glorot11a}
Xavier Glorot, Antoine Bordes, and Yoshua Bengio.
\newblock Deep sparse rectifier neural networks.
\newblock In Geoffrey Gordon, David Dunson, and Miroslav Dudík, editors, {\em Proceedings of the Fourteenth International Conference on Artificial Intelligence and Statistics}, volume~15 of {\em Proceedings of Machine Learning Research}, pages 315--323, Fort Lauderdale, FL, USA, 11--13 Apr 2011. PMLR.

\bibitem{CSI-Extraction}
Francesco Gringoli, Marco Cominelli, Alejandro Blanco, and Joerg Widmer.
\newblock Ax-csi: Enabling csi extraction on commercial 802.11ax wi-fi platforms.
\newblock In {\em Proceedings of the 15th ACM Workshop on Wireless Network Testbeds, Experimental Evaluation CHaracterization}, WiNTECH '21, page 46–53, New York, NY, USA, 2021. Association for Computing Machinery.

\bibitem{7952132}
Shawn Hershey, Sourish Chaudhuri, Daniel P.~W. Ellis, Jort~F. Gemmeke, Aren Jansen, R.~Channing Moore, Manoj Plakal, Devin Platt, Rif~A. Saurous, Bryan Seybold, Malcolm Slaney, Ron~J. Weiss, and Kevin Wilson.
\newblock Cnn architectures for large-scale audio classification.
\newblock In {\em 2017 IEEE International Conference on Acoustics, Speech and Signal Processing (ICASSP)}, pages 131--135, 2017.

\bibitem{huth2004logic}
M~Huth.
\newblock {\em Logic in Computer Science: Modelling and reasoning about systems}.
\newblock Cambridge University Press, 2004.

\bibitem{10.1162/neco_a_01499}
Amirhossein Javanshir, Thanh~Thi Nguyen, M.~A.~Parvez Mahmud, and Abbas~Z. Kouzani.
\newblock {Advancements in Algorithms and Neuromorphic Hardware for Spiking Neural Networks}.
\newblock {\em Neural Computation}, 34(6):1289--1328, 05 2022.

\bibitem{OFDM}
Evgeny Khorov, Anton Kiryanov, Andrey Lyakhov, and Giuseppe Bianchi.
\newblock A tutorial on ieee 802.11ax high efficiency wlans.
\newblock {\em IEEE Communications Surveys Tutorials}, 21(1):197--216, 2019.

\bibitem{kingma2013auto}
Diederik~P Kingma.
\newblock Auto-encoding variational bayes.
\newblock {\em arXiv preprint arXiv:1312.6114}, 2013.

\bibitem{NIPS2012_c399862d}
Alex Krizhevsky, Ilya Sutskever, and Geoffrey~E Hinton.
\newblock Imagenet classification with deep convolutional neural networks.
\newblock In F.~Pereira, C.J. Burges, L.~Bottou, and K.Q. Weinberger, editors, {\em Advances in Neural Information Processing Systems}, volume~25. Curran Associates, Inc., 2012.

\bibitem{lane2010survey}
Nicholas~D Lane, Emiliano Miluzzo, Hong Lu, Daniel Peebles, Tanzeem Choudhury, and Andrew~T Campbell.
\newblock A survey of mobile phone sensing.
\newblock {\em IEEE Communications magazine}, 48(9):140--150, 2010.

\bibitem{726791}
Y.~Lecun, L.~Bottou, Y.~Bengio, and P.~Haffner.
\newblock Gradient-based learning applied to document recognition.
\newblock {\em Proceedings of the IEEE}, 86(11):2278--2324, 1998.

\bibitem{liu2020}
Jenny Liu, Huaizheng Mu, Asad Vakil, Robert Ewing, Xiaoping Shen, Erik Blasch, and Jia Li.
\newblock Human occupancy detection via passive cognitive radio.
\newblock {\em Sensors}, 20(15), 2020.

\bibitem{lv2024efficienteffectivetimeseriesforecasting}
Changze Lv, Yansen Wang, Dongqi Han, Xiaoqing Zheng, Xuanjing Huang, and Dongsheng Li.
\newblock Efficient and effective time-series forecasting with spiking neural networks, 2024.

\bibitem{CSI-Sensing}
Yongsen Ma, Gang Zhou, and Shuangquan Wang.
\newblock Wifi sensing with channel state information: A survey.
\newblock {\em ACM Comput. Surv.}, 52(3), jun 2019.

\bibitem{1997SNN}
Wolfgang Maass.
\newblock Networks of spiking neurons: The third generation of neural network models.
\newblock {\em Neural Networks}, 10(9):1659--1671, 1997.

\bibitem{NEURIPS2018_dc5d637e}
Robin Manhaeve, Sebastijan Dumancic, Angelika Kimmig, Thomas Demeester, and Luc De~Raedt.
\newblock Deepproblog: Neural probabilistic logic programming.
\newblock In S.~Bengio, H.~Wallach, H.~Larochelle, K.~Grauman, N.~Cesa-Bianchi, and R.~Garnett, editors, {\em Advances in Neural Information Processing Systems}, volume~31. Curran Associates, Inc., 2018.

\bibitem{meneghello2022}
Francesca Meneghello, Domenico Garlisi, Nicolò Dal~Fabbro, Ilenia Tinnirello, and Michele Rossi.
\newblock {{SHARP}}: Environment and person independent activity recognition with commodity {{IEEE}} 802.11 access points.
\newblock {\em IEEE Transactions on Mobile Computing}, pages 1--16, 2022.

\bibitem{nair2010rectified}
Vinod Nair and Geoffrey~E Hinton.
\newblock Rectified linear units improve restricted boltzmann machines.
\newblock In {\em Proceedings of the 27th international conference on machine learning (ICML-10)}, pages 807--814, 2010.

\bibitem{9322323}
Takashi Nakamura, Mondher Bouazizi, Kohei Yamamoto, and Tomoaki Ohtsuki.
\newblock Wi-fi-csi-based fall detection by spectrogram analysis with cnn.
\newblock In {\em GLOBECOM 2020 - 2020 IEEE Global Communications Conference}, pages 1--6, 2020.

\bibitem{VideoPose3D}
Dario Pavllo, Christoph Feichtenhofer, David Grangier, and Michael Auli.
\newblock 3d human pose estimation in video with temporal convolutions and semi-supervised training.
\newblock {\em CoRR}, abs/1811.11742, 2018.

\bibitem{renkens2014explanation}
Joris Renkens, Angelika Kimmig, Guy Van~den Broeck, and Luc De~Raedt.
\newblock Explanation-based approximate weighted model counting for probabilistic logics.
\newblock In {\em Proceedings of the AAAI Conference on Artificial Intelligence}, volume~28, 2014.

\bibitem{rosenblatt1958perceptron}
Frank Rosenblatt.
\newblock The perceptron: a probabilistic model for information storage and organization in the brain.
\newblock {\em Psychological review}, 65(6):386, 1958.

\bibitem{roy2019towards}
Kaushik Roy, Akhilesh Jaiswal, and Priyadarshini Panda.
\newblock Towards spike-based machine intelligence with neuromorphic computing.
\newblock {\em Nature}, 575(7784):607--617, 2019.

\bibitem{pmlr-v84-runge18a}
Jakob Runge.
\newblock Conditional independence testing based on a nearest-neighbor estimator of conditional mutual information.
\newblock In Amos Storkey and Fernando Perez-Cruz, editors, {\em Proceedings of the Twenty-First International Conference on Artificial Intelligence and Statistics}, volume~84 of {\em Proceedings of Machine Learning Research}, pages 938--947. PMLR, 09--11 Apr 2018.

\bibitem{runge2023causal}
Jakob Runge, Andreas Gerhardus, Gherardo Varando, Veronika Eyring, and Gustau Camps-Valls.
\newblock Causal inference for time series.
\newblock {\em Nature Reviews Earth Environment}, 4(7):487--505, 2023.

\bibitem{doi:10.1126/sciadv.aau4996}
Jakob Runge, Peer Nowack, Marlene Kretschmer, Seth Flaxman, and Dino Sejdinovic.
\newblock Detecting and quantifying causal associations in large nonlinear time series datasets.
\newblock {\em Science Advances}, 5(11):eaau4996, 2019.

\bibitem{sang2005performing}
Tian Sang, Paul Beame, and Henry~A Kautz.
\newblock Performing bayesian inference by weighted model counting.
\newblock In {\em AAAI}, volume~5, pages 475--481, 2005.

\bibitem{1467551}
T.~Serre, L.~Wolf, and T.~Poggio.
\newblock Object recognition with features inspired by visual cortex.
\newblock In {\em 2005 IEEE Computer Society Conference on Computer Vision and Pattern Recognition (CVPR'05)}, volume~2, pages 994--1000 vol. 2, 2005.

\bibitem{storf2009rule}
Holger Storf, Martin Becker, and Martin Riedl.
\newblock Rule-based activity recognition framework: Challenges, technique and learning.
\newblock In {\em 3rd International Conference on Pervasive Computing Technologies for Healthcare}, pages 1--7. IEEE, 2009.

\bibitem{theekakul2011rule}
Pitchakan Theekakul, Surapa Thiemjarus, Ekawit Nantajeewarawat, Thepchai Supnithi, and Kaoru Hirota.
\newblock A rule-based approach to activity recognition.
\newblock In {\em Proceedings of the 5th International Conference on Knowledge, Information, and Creativity Support Systems}, pages 204--215. Springer-Verlag, 2010.

\bibitem{venema2017temporal}
Yde Venema.
\newblock Temporal logic.
\newblock {\em The Blackwell guide to philosophical logic}, pages 203--223, 2017.

\bibitem{vlasselaer2014compiling}
Jonas Vlasselaer, Joris Renkens, Guy Van~den Broeck, and Luc De~Raedt.
\newblock Compiling probabilistic logic programs into sentential decision diagrams.
\newblock In {\em Workshop on Probabilistic Logic Programming (PLP), Vienna}, 2014.

\bibitem{8735849}
Zhengjie Wang, Kangkang Jiang, Yushan Hou, Zehua Huang, Wenwen Dou, Chengming Zhang, and Yinjing Guo.
\newblock A survey on csi-based human behavior recognition in through-the-wall scenario.
\newblock {\em IEEE Access}, 7:78772--78793, 2019.

\bibitem{brainsci12070863}
Kashu Yamazaki, Viet-Khoa Vo-Ho, Darshan Bulsara, and Ngan Le.
\newblock Spiking neural networks and their applications: A review.
\newblock {\em Brain Sciences}, 12(7), 2022.

\end{thebibliography}

\appendix

\chapter{Bayesian Hypothesis Testing for Classifier Comparison}
\label{app:BHT}

In this section, we outline the Bayesian hypothesis testing approach used to determine whether two classifiers are statistically equivalent or different in their performance across a multiclass classification problem. This method compares two classifiers, \( A \) and \( B \), using confusion matrices and assesses whether the observed counts suggest that their classification abilities are drawn from the same or different distributions. The key components of the test include multinomial models, Dirichlet priors, and Bayes factors.

\section{Problem Definition}
We denote the confusion matrices for classifiers \( A \) and \( B \) as \( \mathbf{C_A} \) and \( \mathbf{C_B} \), both of size \( 7 \times 7 \). Let \( \mathbf{c_A} \) and \( \mathbf{c_B} \) represent the set of counts associated with correct classifications and errors for classifiers \( A \) and \( B \), respectively:

\begin{itemize}
    \item \( \mathbf{c_A} = [c_1, c_2, \ldots, c_7, e_1, e_2, \ldots, e_7] \)
    \item \( c_i \) corresponds to the count of correct classifications for class \( i \) (diagonal elements of the confusion matrix).
    \item \( e_i \) represents the total number of misclassifications for true class \( i \) (sum of off-diagonal elements for each row).
\end{itemize}

The goal of this Bayesian analysis is to decide between two competing hypotheses:
\begin{itemize}
    \item \textbf{\( H_{\text{same}} \)}: The two classifiers are essentially performing the same, i.e., their results are drawn from the same multinomial distribution.
    \item \textbf{\( H_{\text{indep}} \)}: The classifiers are different, i.e., their results are drawn from different multinomial distributions.
\end{itemize}

\section{Dirichlet-Multinomial Model}
We assume that the observed counts \( \mathbf{c_A} \) and \( \mathbf{c_B} \) follow multinomial distributions, parameterized by probability vectors \( \boldsymbol{\alpha} \) and \( \boldsymbol{\beta} \). For each classifier, the probability of a specific outcome vector is modeled by a multinomial likelihood. The prior over these probability vectors is chosen to be a Dirichlet distribution, which is conjugate to the multinomial, facilitating analytical tractability in Bayesian analysis.

\section{Likelihood Functions}
For classifier \( A \):
\begin{equation}
    p(\mathbf{c_A} \mid \boldsymbol{\alpha}) = \prod_{i=1}^{14} \alpha_i^{c_{A, i}}
\end{equation}
where \( \alpha_i \) represents the probability of outcome \( i \) for classifier \( A \), and \( c_{A, i} \) is the count for that outcome.

For classifier \( B \):
\begin{equation}
    p(\mathbf{c_B} \mid \boldsymbol{\beta}) = \prod_{i=1}^{14} \beta_i^{c_{B, i}}
\end{equation}

\section{Prior Distribution}
The Dirichlet prior is parameterized as:
\begin{equation}
    p(\boldsymbol{\alpha}) = \frac{1}{Z(\mathbf{u})} \prod_{i=1}^{14} \alpha_i^{u_i - 1}
\end{equation}
where \( Z(\mathbf{u}) \) is the normalization constant of the Dirichlet distribution, and \( \mathbf{u} \) is a vector that assigns higher prior mass to correct classifications and near-zero values to errors:
\begin{itemize}
    \item For correct classifications (\( c_1, c_2, \ldots, c_7 \)), we set \( u_i = u_{\text{correct}} \), which could be 1 for a uniform distribution.
    \item For misclassifications (\( e_1, e_2, \ldots, e_7 \)), we set \( u_i = u_{\text{error}} \), with \( u_{\text{error}} \ll 1 \) (e.g., 0.01), indicating a near-zero prior for errors.
\end{itemize}

\section{Evidence Calculation}
The evidence for each hypothesis is calculated by integrating out the parameters \( \boldsymbol{\alpha} \) and \( \boldsymbol{\beta} \).

\subsection{For \( H_{\text{indep}} \)}
The hypothesis \( H_{\text{indep}} \) assumes that \( \boldsymbol{\alpha} \) and \( \boldsymbol{\beta} \) are independent:
\begin{equation}
    p(\mathbf{c_A}, \mathbf{c_B} \mid H_{\text{indep}}) = \int p(\mathbf{c_A} \mid \boldsymbol{\alpha}) p(\boldsymbol{\alpha}) d\boldsymbol{\alpha} \int p(\mathbf{c_B} \mid \boldsymbol{\beta}) p(\boldsymbol{\beta}) d\boldsymbol{\beta}
\end{equation}
Using the Dirichlet-multinomial conjugacy, the above integrals simplify to:
\begin{equation}
    p(\mathbf{c_A}, \mathbf{c_B} \mid H_{\text{indep}}) = \frac{Z(\mathbf{u} + \mathbf{c_A}) Z(\mathbf{u} + \mathbf{c_B})}{Z(\mathbf{u})^2}
\end{equation}

\subsection{For \( H_{\text{same}} \)}
The hypothesis \( H_{\text{same}} \) assumes that both classifiers share the same parameter vector \( \boldsymbol{\alpha} \):
\begin{equation}
    p(\mathbf{c_A}, \mathbf{c_B} \mid H_{\text{same}}) = \int p(\mathbf{c_A} \mid \boldsymbol{\alpha}) p(\mathbf{c_B} \mid \boldsymbol{\alpha}) p(\boldsymbol{\alpha}) d\boldsymbol{\alpha}
\end{equation}
which simplifies to:
\begin{equation}
    p(\mathbf{c_A}, \mathbf{c_B} \mid H_{\text{same}}) = \frac{Z(\mathbf{u} + \mathbf{c_A} + \mathbf{c_B})}{Z(\mathbf{u})}
\end{equation}

\subsection{Bayes Factor}
The Bayes factor \( BF \) is used to compare the evidence for \( H_{\text{indep}} \) to that of \( H_{\text{same}} \):
\begin{equation}
    BF = \frac{p(\mathbf{c_A}, \mathbf{c_B} \mid H_{\text{indep}})}{p(\mathbf{c_A}, \mathbf{c_B} \mid H_{\text{same}})}
\end{equation}
\begin{equation}
    BF = \frac{Z(\mathbf{u} + \mathbf{c_A}) Z(\mathbf{u} + \mathbf{c_B})}{Z(\mathbf{u}) Z(\mathbf{u} + \mathbf{c_A} + \mathbf{c_B})}
\end{equation}
This factor indicates the strength of evidence in favor of the classifiers being different (\( H_{\text{indep}} \)) versus them being equivalent (\( H_{\text{same}} \)). If \( BF > 1 \), there is evidence that the classifiers perform differently, while \( BF < 1 \) suggests that they are essentially performing the same.

\section{Interpretation}
\begin{itemize}
    \item \textbf{\( BF > 1 \)}: Strong evidence that classifiers \( A \) and \( B \) are statistically different, implying that they may be suited for different applications or show divergent performance in certain classes.
    \item \textbf{\( BF < 1 \)}: Strong evidence that the classifiers are statistically equivalent, which might indicate that they perform similarly across all classes.
\end{itemize}

This Bayesian hypothesis testing method provides a rigorous probabilistic framework to assess whether two classifiers are performing significantly differently or not, leveraging the flexibility of the Dirichlet-multinomial model to accommodate class-wise variation in performance. The use of priors that favor correct classifications over errors provides further insight into the practical efficacy of each classifier.

\end{document}